\documentclass{cja}

\usepackage{xurl}
\usepackage{bm}
\usepackage{threeparttable}
\usepackage{cases}
\usepackage{booktabs}
\usepackage{graphicx}
\usepackage{tabularx}
\usepackage{amsmath}      
\usepackage{algorithm}    
\usepackage{algorithmic}  
\usepackage{multirow}     
\usepackage{enumitem}
\usepackage{siunitx}
\usepackage{threeparttable}
\usepackage{graphicx}
\usepackage{caption}
\usepackage{eso-pic}
\usepackage{hyperref}
\usepackage{cleveref}

\begin{document}



\title{Object-aware graph matching network for cross-domain remote sensing image localization}




\author[a]{Tao LIU}

\author[a,b]{Kan REN\corref{cor}}
\ead{k.ren@njust.edu.cn}

\author[a,b]{Qian CHEN}

\shortauthors{T. LIU, K. REN and Q. CHEN}

\cortext[cor]{Corresponding author}

\affiliation[a]{
  organization = {Jiangsu Key Laboratory of Visual Sensing and Intelligent Perception, Nanjing University of Science and Technology},
  addressline = {Nanjing 210094},
  country = {China}
}

\affiliation[b]{
  organization = {National Key Laboratory of Dynamic Measurement Technology and Instrumentation for Extreme Environments, Nanjing University of Science and Technology},
  addressline = {Nanjing 210094},
  country = {China}
}

\begin{abstract}
  Cross-domain and cross-modal remote sensing image geo-localization remains challenging due to large appearance discrepancies and unstable semantic correspondence across heterogeneous sensors and platforms. Existing methods mainly rely on scene-level or global representations, which often struggle to achieve reliable alignment in complex environments, especially under severe modality gaps such as infrared-to-visible matching. To facilitate research in this setting, this study introduces IRVL328, a new infrared-visible remote sensing localization dataset designed to reflect challenging cross-modal variations. Meanwhile, this study proposes an object-aware graph matching framework that integrates object detection with dual-graph neural reasoning, where salient structural regions are represented as graph nodes and both inter-image correspondences and intra-image relations are jointly modeled; a training-only node alignment strategy is further introduced to enhance supervision without increasing inference complexity. Experimental results show that the proposed method achieves competitive performance on SUES-200 and strong performance on IRVL328 and DenseUAV, particularly in challenging cross-modal settings.
\end{abstract}



\frontheader{Chinese Journal of Aeronautics, (year), \textbf{volum}(number): page--page}
\articledoi{10.1016/j.cja.2026.xx.xxx}
\frontfooter{%
  1000-9361 © 2026 Chinese Society of Aeronautics and Astronautics.
  Production and hosting by Elsevier Ltd.\par
  This is an open access article under the CC BY-NC-ND license
  (\url{http://creativecommons.org/licenses/by-nc-nd/4.0/}).
}
\received{5 February 2026}
\revised{16 May 2026}
\accepted{6 July 2026}

\AddToShipoutPictureFG*{%
  \AtPageUpperLeft{%
    \raisebox{-1.0cm}{\hspace*{1.6cm}\textbf{Final Accepted Version}}%
  }%
}

\maketitle


\section{Introduction}
\label{sec:introduction}

Recent advances in remote sensing platforms have enabled multi-source Earth observation using satellites, aerial photography, and unmanned aerial vehicles, providing rich support for geospatial understanding and localization.~\cite{ref1} Compared with Global Navigation Satellite System (GNSS)-based localization, which may suffer from signal blockage and electromagnetic interference in complex environments,~\cite{ref2} vision-based geo-localization offers an alternative by matching query images against geo-referenced databases.~\cite{ref3,ref4,ref5} However, cross-domain remote sensing localization remains highly challenging because images acquired from different platforms, times, and sensing modalities often exhibit substantial discrepancies.

\begin{figure*}[t]
    \centering
    \includegraphics[width=0.9\linewidth]{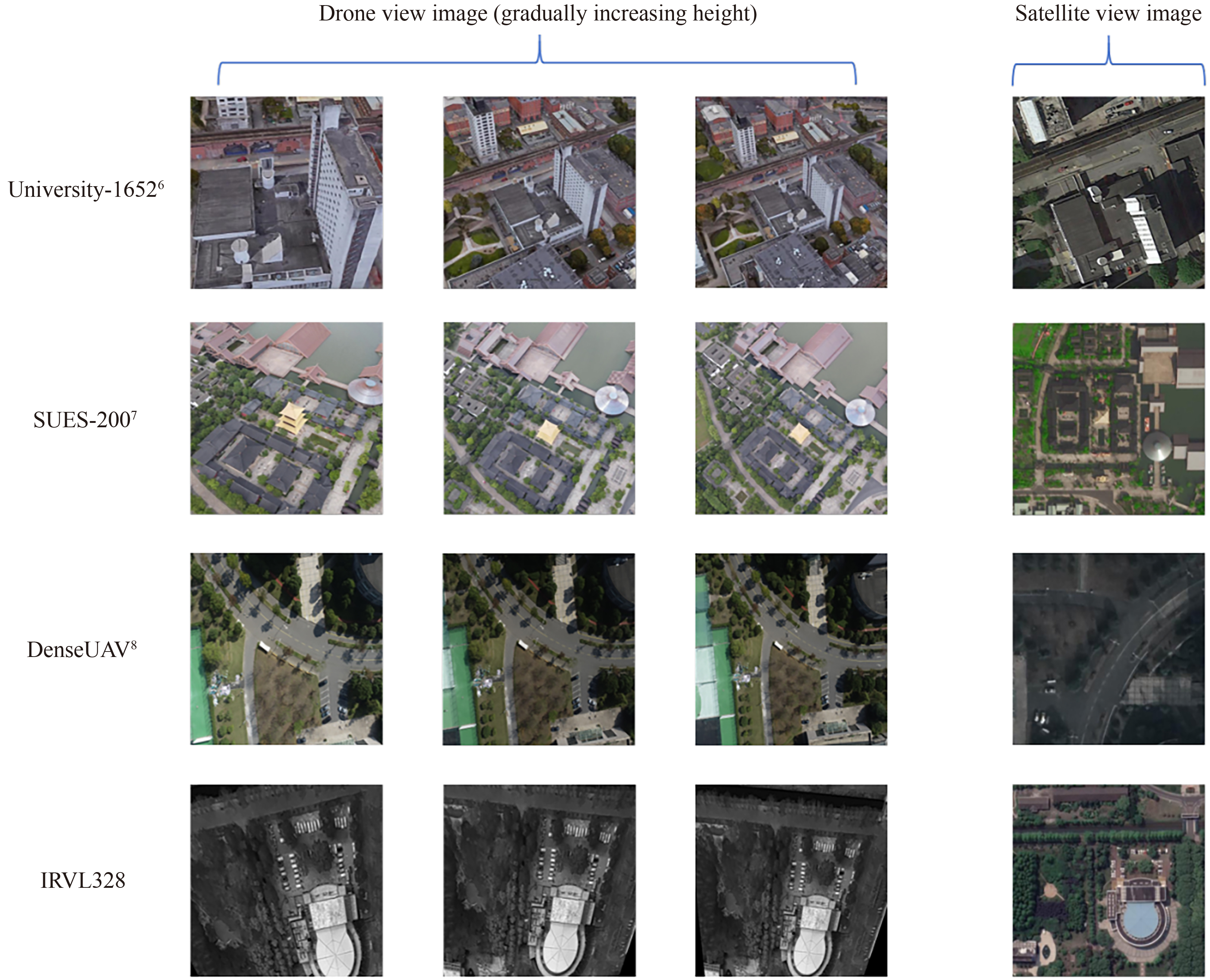}
    \caption{Representative datasets utilized in this study.}
    \label{fig1}
\end{figure*}

As illustrated in Fig.~\ref{fig1}, the difficulty of cross-view geo-localization mainly stems from three types of domain gaps, including viewpoint gaps, temporal gaps, and modality gaps. Representative datasets employed in this study are presented, including the virtual dataset University-1652,~\cite{ref19} the real-world dataset SUES-200,~\cite{ref20} the DenseUAV dataset,~\cite{ref21} and the self-collected infrared/visible drone images with their corresponding satellite imagery. These datasets involve various domain variations, such as different viewpoints, seasonal vegetation changes, illumination and shadow variations, transient objects (e.g., vehicles), and radiometric/geometric discrepancies caused by heterogeneous sensors. Specifically, the viewpoint gap is pronounced because drone images are usually captured from oblique and low-altitude perspectives, whereas satellite images are typically acquired from nadir views at much higher altitudes, resulting in significant geometric and perspective differences. The temporal gap caused by seasonal changes, illumination variations, shadows, and transient objects further reduces the appearance consistency across images. Moreover, the modality gap becomes more challenging in infrared-visible scenarios, where images exhibit substantial differences in radiometric characteristics and visual appearance.~\cite{xiao2023long} These factors make reliable cross-platform matching under varying sensing conditions extremely challenging. Therefore, this work aims to bridge these domain gaps and achieve accurate cross-view geo-localization.

Existing cross-view geo-localization methods mainly rely on Convolutional Neural Networks (CNNs)~\cite{ref13} or Transformers~\cite{ref14} to learn global image representations for retrieval.~\cite{ref15,ref16} Although effective in many settings, global feature learning often entangles stable structural cues with unstable appearance patterns, making it difficult to preserve reliable correspondences under large viewpoint, temporal, and modality variations. Local correspondence or keypoint-style matching can partially improve spatial sensitivity, but is often unstable when object appearance, scale, and imaging modality change significantly. Earlier attempts based on geometric transformation~\cite{ref6} or generative alignment~\cite{ref7,ref8,ref9} also remain insufficient for complex real-world scenarios, as they may require strong prior assumptions or introduce artifacts and optimization instability. Recent studies have explored multimodal remote sensing visual localization~\cite{2025CDM, ji2025mmgeo, yin2026triple} combining text and images, yet the field of infrared-visible visual localization~\cite{xiao2023long} still lacks paired datasets. As a result, existing methods still struggle to robustly align semantically corresponding regions across heterogeneous remote sensing images.

In remote sensing geo-localization, stable objects and their relative spatial organization are often more transferable across viewpoints, time, and modalities than holistic texture or scene appearance. Structural landmarks such as buildings, bridges, and road junctions usually provide stronger localization cues than globally aggregated features, while their relationships offer complementary context for disambiguation. This observation motivates an object-aware structured matching strategy: instead of relying solely on scene-level descriptors, the model should explicitly represent salient objects and reason over their spatial and semantic relations. Graph representation is well suited for this purpose, because it naturally models object instances as nodes and their interactions as edges, enabling structured correspondence learning beyond unordered feature aggregation.

Based on this motivation, this study proposes an Object-aware Graph Matching Network (OGMN) for cross-domain remote sensing geo-localization. OGMN first detects salient structural objects from drone and satellite images and converts them into graph-based representations that encode both semantic attributes and spatial relations. It then employs graph neural reasoning to model intra-image structure and enhance feature abstraction under domain discrepancy. In addition, a training-only node alignment mechanism is introduced to strengthen inter-image correspondence supervision, while being removed during inference to preserve efficiency.

The main contributions of this work are summarized as follows:
\begin{enumerate}[label=(\arabic*), leftmargin=*, itemsep=0pt, topsep=2pt, parsep=0pt]
    \item This study proposes an object-aware graph-based matching framework for cross-domain remote sensing geo-localization, which explicitly represents salient objects and their structural relations for more robust matching across viewpoints, time, and sensing modalities.
    \item This study introduces a training-only node alignment strategy that strengthens correspondence supervision during training while preserving efficient inference.
    \item This study introduces IRVL328, a new infrared-visible drone-satellite dataset for cross-modal geo-localization, to support future research on nighttime and heterogeneous sensing scenarios.
\end{enumerate}

\section{Related work}

\subsection{Image geo-localization}
Cross-view image geo-localization is commonly formulated as an image retrieval problem, where the goal is to learn viewpoint-invariant representations that bring images from the same location closer while separating those from different locations in a shared feature space. Early methods mainly relied on dual-branch CNN architectures to learn such representations under supervised metric learning. Although effective when viewpoint differences are moderate, these models often degrade under large spatial or appearance discrepancies.

To reduce the view gap, geometry-based transformations such as polar coordinate conversion~\cite{ref6} and perspective warping~\cite{ref7} have been explored. However, these methods usually depend on relatively accurate prior alignment and may introduce information loss during transformation. Generative Adversarial Network (GAN)-based approaches~\cite{ref8,ref9} further attempt to translate images across views, but the generated results often deviate from real image distributions, which limits generalization in complex scenes.

More recently, Transformer-based models have shown strong performance in this area. For example, L2LTR~\cite{ref22} combines ResNet~\cite{ref71} and Vision Transformer (ViT)~\cite{ref70} to incorporate both convolutional inductive bias and global context modeling. Subsequent works~\cite{ref23,ref24,ref25,ref26} introduce operations such as region cropping and semantic segmentation to improve fine-grained matching. TirSA~\cite{ref27} leverages self-supervised learning to suppress irrelevant background interference. Attention-based models such as MLPCAN~\cite{ref28} and SDPL~\cite{ref29} enhance discriminative feature learning through multi-level attention and shift-based fusion. MCCG~\cite{ref30} further extends CNN-based retrieval using ConvNeXt,~\cite{ref31} while Sample4Geo~\cite{ref32} achieves strong performance with refined contrastive learning strategies.~\cite{ref33}

Although these retrieval-based methods are effective when the viewpoint gap is moderate and the scene appearance remains relatively consistent, they still rely heavily on holistic visual representations. In cross-domain remote sensing geo-localization, however, large viewpoint changes, seasonal variation, and modality discrepancies can substantially distort global appearance and weaken reliable correspondence. This limitation motivates representations that focus on stable localization cues and their structural organization, rather than depending solely on scene-level texture.

\subsection{Image matching}
Image matching aims to establish correspondences between images captured under different viewpoints, times, or sensing modalities. Classical methods based on handcrafted features, such as SIFT,~\cite{ref34} detect local keypoints and describe them with invariant descriptors. While these methods are effective in controlled settings, they often become fragile under severe illumination changes, large viewpoint differences, or strong perspective distortion.

\begin{figure}[t]
    \centering
    \includegraphics[width=0.9\linewidth]{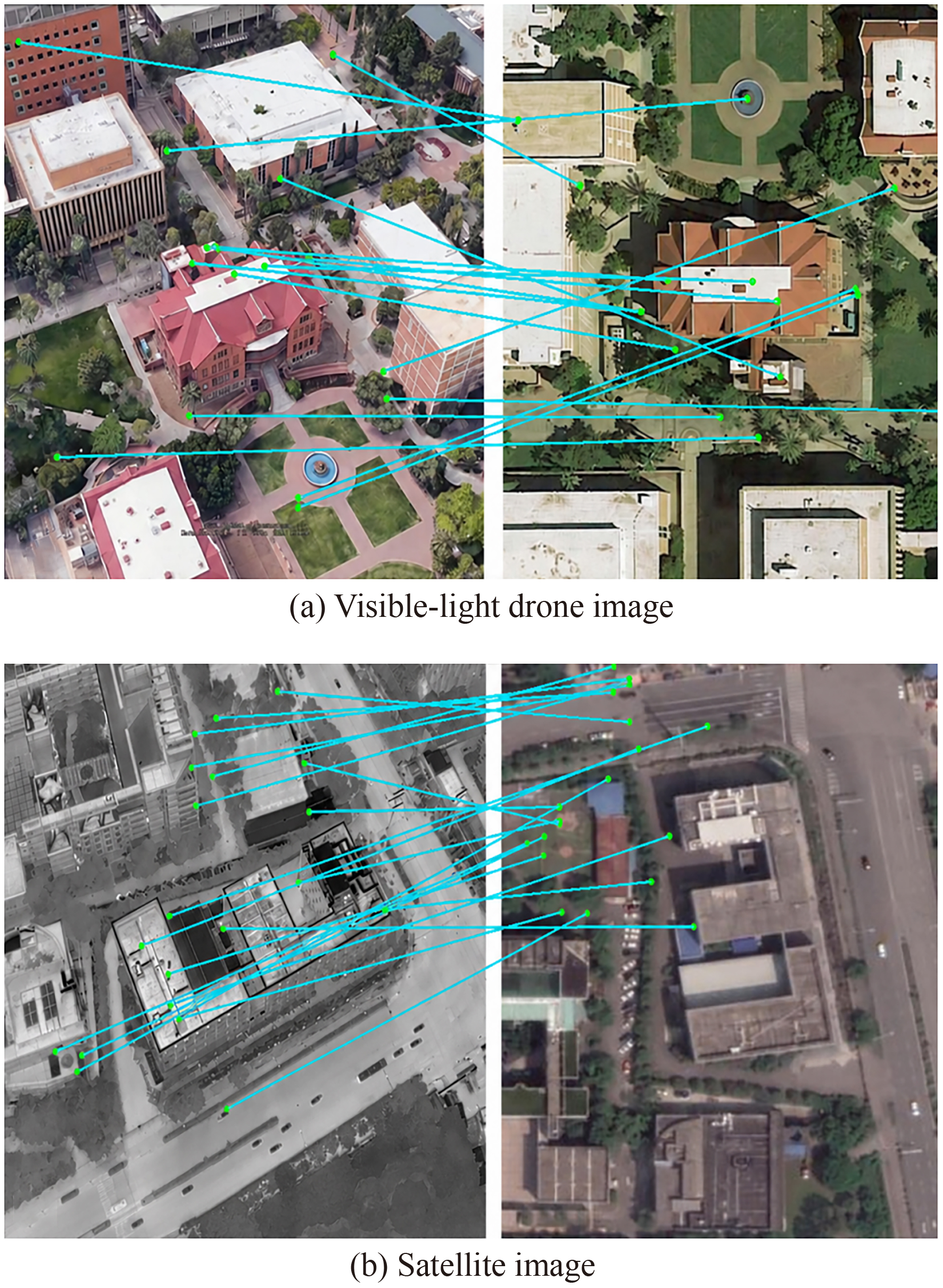}
    \caption{Cross-view image matching using LightGlue.}
    \label{fig2}
\end{figure}

Deep learning has significantly advanced image matching in recent years. SuperGlue~\cite{ref35} combines graph-style feature interaction with Transformer encoders and optimal transport, enabling high-precision local correspondence estimation. LightGlue~\cite{ref36} further improves the balance between efficiency and accuracy through dynamic adaptation. These methods have demonstrated strong performance in many visual matching scenarios.

Nevertheless, existing methods still face significant challenges in cross-domain scenarios, such as drone-to-satellite alignment, where large scale variations, severe viewpoint changes, and information loss substantially degrade the reliability of local correspondences. Specifically, two representative challenging cases are considered: Fig.~\ref{fig2}(a) visible-light drone and satellite images, where significant viewpoint and scale differences result in numerous mismatched local features; and Fig.~\ref{fig2}(b) infrared and visible-light images, where severe viewpoint variations and radiometric appearance discrepancies make conventional feature matching completely fail. These issues become even more pronounced in infrared-visible settings, where consistent low-level keypoints are often sparse or unavailable. Therefore, robust cross-domain remote sensing localization requires a higher-level representation that focuses on semantically meaningful and structurally stable regions, rather than relying solely on dense local correspondences.

\subsection{Object detection}
Object detection is one of the fundamental tasks in computer vision, and deep learning has substantially improved its performance. Two-stage detectors such as Faster R-CNN~\cite{ref37} generally emphasize detection accuracy, whereas one-stage frameworks such as the YOLO series~\cite{ref38} are designed for higher efficiency and real-time deployment.

In aerial and remote sensing imagery, existing detection studies have mainly focused on targets such as aircraft, ships, and vehicles, as well as certain large stationary objects including bridges, oil fields, and water bodies.~\cite{ref39,ref40} However, datasets specifically emphasizing spatio-temporally stable landmarks that are useful for geo-localization, such as buildings, road intersections, and playgrounds, remain limited. To address this issue, this study integrates and re-annotates multiple drone localization datasets, including University-1652, SUES-200, and DenseUAV, to provide object-level labels that are more relevant to cross-domain geo-localization.

Unlike conventional aerial detection tasks,  the objective of this study is not object recognition for its own sake. Existing aerial detection methods are primarily designed for surveillance or category-level detection, and the selected object categories are not always aligned with landmarks that remain stable and discriminative across viewpoints, time, and sensing modalities. In the proposed framework, object detection serves as a task-driven front-end for identifying structurally stable landmarks, which subsequently support more reliable cross-domain matching.

\subsection{Graph matching}
Graphs provide a flexible representation for structured data by modeling entities as nodes and their interactions as edges. Graph matching measures the similarity between two graphs and has traditionally been formulated as an NP-hard optimization problem.~\cite{ref42} Although classical methods can solve this problem under certain constraints, computational scalability remains a major challenge.

Recent advances in deep learning have revived interest in graph matching and graph similarity learning. Early neural graph matching methods~\cite{ref43} introduced learnable node and edge embeddings, while attention-based mechanisms~\cite{ref44} further improved matching efficiency and representation capability. SimGNN~\cite{ref45} combined graph embedding with neural similarity estimation to enable scalable graph comparison. Beyond pure vision tasks, graph-based matching has also shown promising performance in multimodal applications such as image-text retrieval.~\cite{ref46}

These developments highlight the potential of graph-based reasoning for structured correspondence learning. However, most existing graph matching methods are not designed for cross-domain remote sensing geo-localization, where the central challenge is to align temporally stable landmarks under large viewpoint, temporal, and modality discrepancies. Different from generic graph comparison or semantic alignment tasks, this study focuses on retrieval-oriented geo-localization based on object-level semantic cues and their spatial organization. This motivates a task-specific graph matching framework tailored to remote sensing imagery rather than a direct adaptation of existing graph-matching pipelines.

Overall, existing studies have made substantial progress in geo-localization, image matching, object detection, and graph-based reasoning. However, they are still insufficient for cross-domain remote sensing localization, where reliable matching must simultaneously address viewpoint gap, temporal variation, and modality discrepancy. In particular, existing approaches often rely on holistic scene representations, unstable local correspondences, or generic graph constructions that are not specifically designed for stable localization landmarks. In contrast, the proposed approach is motivated by the observation that structurally stable objects and their spatial organization provide more transferable cues for geo-localization. Based on this insight, this study develops a task-specific object-aware graph matching framework with training-only node alignment, aiming to improve robustness without increasing inference complexity.

\section{Proposed method}
\begin{figure*}[t]
    \centering
    \includegraphics[width=0.95 \linewidth]{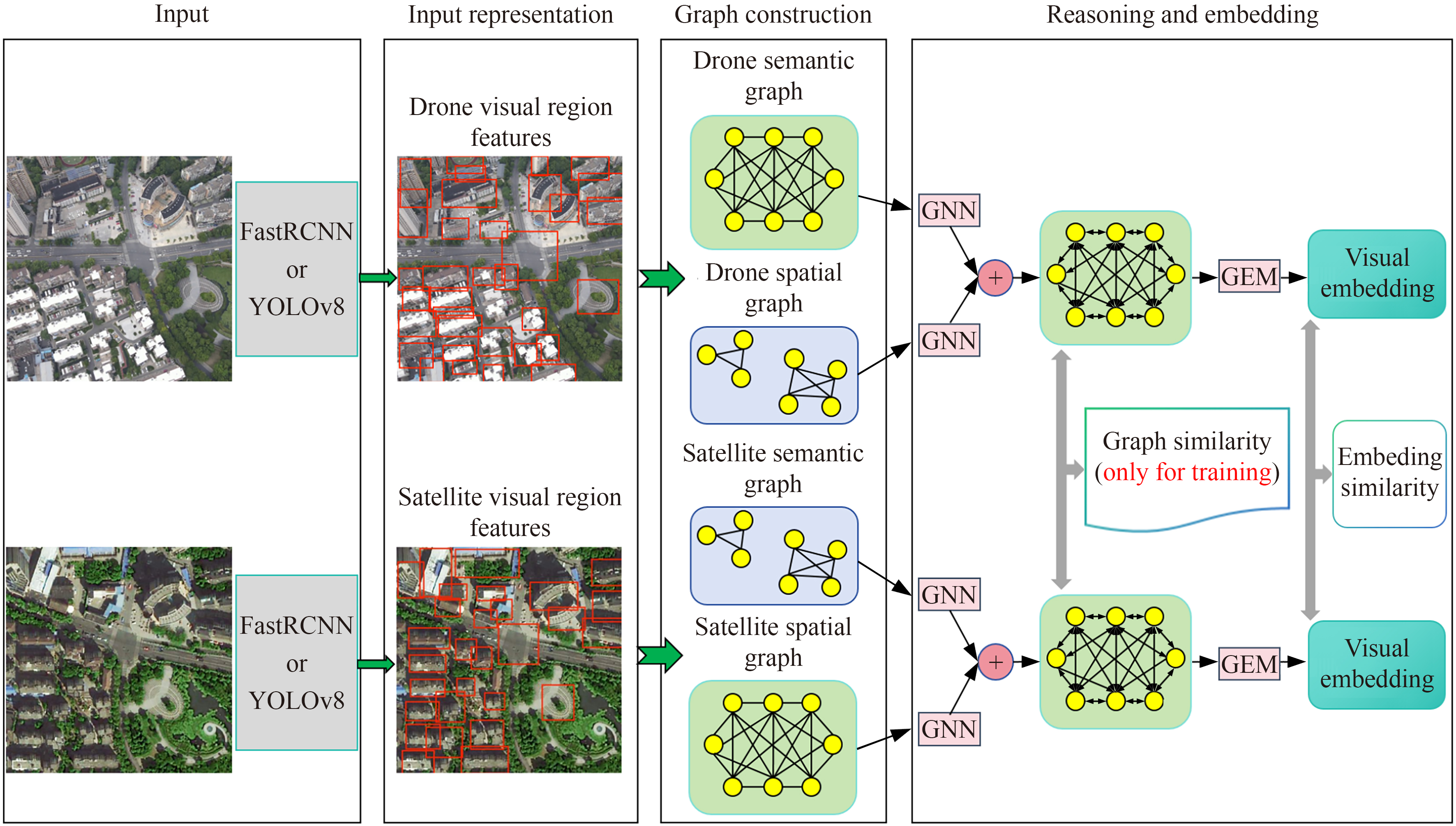}
    \caption{Overview of proposed matching framework.}
    \label{fig3}
\end{figure*}

The workflow of the proposed model is illustrated in Fig.~\ref{fig3}. The framework consists of four main components: (A) salient region extraction, (B) drone-satellite graph construction, (C) graph-based feature reasoning, and (D) training strategy and loss optimization. 

Specifically, in component (A), salient regions are first detected from drone and satellite images using object detectors (e.g., Faster R-CNN or YOLOv8) with top-down attention, which aims to extract discriminative semantic regions while suppressing irrelevant background information. In component (B), visual graphs are constructed for both drone and satellite images by incorporating spatial layouts and semantic relationships among the extracted regions. In component (C), a Graph Neural Network (GNN) is employed to capture intra-graph and inter-graph relationships, enabling effective structural reasoning and generating graph-level embeddings for cross-view matching. Finally, component (D) introduces the training strategy and loss function design, where both graph node similarity and embedding similarity objectives are optimized. Notably, node similarity is only utilized during training to enhance feature learning, while inference relies solely on the generated embeddings to maintain efficient matching.  

\subsection{Object detection module}
\label{III-A}
This study adopts two representative object detection algorithms: the classic two-stage Faster R-CNN~\cite{ref37} and the mainstream YOLOv8,~\cite{ref47} to detect salient target regions in images.  

The choice of detection categories is critical for drone and satellite applications in aerial remote sensing. Most categories in general datasets such as PASCAL VOC~\cite{ref48} and MS COCO~\cite{ref49} are either absent or imprecisely defined in aerial imagery, while existing aerial detection datasets mainly focus on dynamic moving targets, which are unsuitable for visual localization. Therefore, it is necessary to redefine categories that are stable across time and space and applicable to both drone and satellite images.  

\begin{table}[t]
  \captionsetup{width=\linewidth}
  \centering
  \caption{Categories of aerial targets considered in this study.}
  \label{tab:table1}

  \scriptsize
  \setlength{\tabcolsep}{4pt}
  \renewcommand{\arraystretch}{1.10}

  \begin{tabularx}{\linewidth}
    {>{\centering\arraybackslash}p{0.10\linewidth}
     >{\raggedright\arraybackslash}X
     >{\centering\arraybackslash}p{0.10\linewidth}
     >{\raggedright\arraybackslash}X}

    \toprule

    No. of targets & Category & No. of targets & Category \\

    \midrule

    1  & Low-rise residential  & 9  & Irregular buildings \\
    2  & Mid-rise residential  & 10 & Intersection \\
    3  & High-rise residential & 11 & Parking lot \\
    4  & Saving box            & 12 & Chimney \\
    5  & Baseball field        & 13 & Tennis court \\
    6  & Basketball field      & 14 & Football field \\
    7  & Playground            & 15 & Rugby field \\
    8  & Bridge                & 16 & Lighthouse \\

    \bottomrule
  \end{tabularx}

\end{table}

Considering the characteristics of image matching tasks, this study aims to identify long-term localization-relevant targets common to both modalities, while excluding categories sensitive to temporal and seasonal variations. Based on prior research~\cite{ref50,ref51} and practical analysis, this study selects 16 categories as key targets for image matching. Specifically, Low-rise residential denotes buildings up to three stories, Mid-rise residential refers to buildings of four to nine stories, and High-rise residential refers to buildings of ten or more stories. Details are summarized in Table~\ref{tab:table1}.  

These 16 categories are selected because they satisfy three important requirements for cross-view geo-localization.~\cite{xue2023sfd2, gao2025semantic} First, they are generally stable man-made structures or functional regions that persist over long time spans and remain observable across different imaging dates, which makes them more reliable under temporal change. Second, compared with dynamic or transient objects such as vehicles and pedestrians, these targets are less sensitive to seasonal variation, traffic conditions, and short-term scene changes, and are therefore more suitable for localization-oriented matching. Third, these categories usually exhibit relatively clear spatial layouts and semantic meanings, which provide effective support for the subsequent graph construction and relational modeling process.

Existing studies have explored object detection in aerial imagery, yet a dedicated and systematic comparison of detector choices for remote sensing image–based visual retrieval and localization is still relatively scarce. Accordingly, this study conducts comparative experiments using two representative architectures, Faster R-CNN and YOLOv8. The implementation pipeline and technical specifications are provided in Section~\ref{V-C}. 

\subsection{Drone and satellite image representation}

Given a drone image \( \bm{U} \), this study applies the object detection methods to extract \( m \) salient regions. The global feature map of the drone image, \( \bm{r}_{\mathrm{u}}^{(0)} \), is concatenated with the feature maps of the \( m \) bounding boxes, \( \{\bm{r}_{\mathrm{u}}^{(i)} \mid i=1,2,\dots,m\} \), yielding the node representation of the drone image.  

\begin{equation}
  \bm{R}_{\mathrm{u}} = \{\bm{r}_{\mathrm{u}}^{(i)} \mid i=0,1,\dots,m, \bm{r}_{\mathrm{u}}^{(i)} \in \mathbb{R}^D\}
\end{equation}
Similarly, the satellite image is represented as  

\begin{equation}
  \bm{R}_{\mathrm{s}} = \{\bm{r}_{\mathrm{s}}^{(j)} \mid j=0,1,\dots,n, \bm{r}_{\mathrm{s}}^{(j)} \in \mathbb{R}^D\}
\end{equation}

A fully connected layer of dimension \( D \) encodes these features into \( D \)-dimensional vectors, as shown in Eqs.~\eqref{eq:drone_feat} and \eqref{eq:sat_feat}  

\begin{equation}
  \bm{v}_{\mathrm{u}}^{(i)} = \bm{W}_\mathrm{f} \bm{r}_{\mathrm{u}}^{(i)} + \bm{b}_\mathrm{f}
  \label{eq:drone_feat}
\end{equation}

\begin{equation}
  \bm{v}_{\mathrm{s}}^{(j)} = \bm{W}_\mathrm{f} \bm{r}_{\mathrm{s}}^{(j)} + \bm{b}_\mathrm{f}
  \label{eq:sat_feat}
\end{equation} where \( \bm{W}_\mathrm{f} \) and \( \bm{b}_\mathrm{f} \) denote the weight and bias parameters of the fully connected layer.

Thus, the node sets are obtained as  
\(\bm{V} = \{\bm{V}_{\mathrm{u}}, \bm{V}_{\mathrm{s}}\}\),  
with \(\bm{V}_{\mathrm{u}} = \{\bm{v}_{\mathrm{u}}^{(i)} \mid i=0,1,\dots,m\}\) for the drone view and 
\(\bm{V}_{\mathrm{s}} = \{\bm{v}_{\mathrm{s}}^{(j)} \mid j=0,1,\dots,n\}\) for the satellite view.  

Here, the node indexed by \(i=0\) or \(j=0\) denotes the global node, while the remaining nodes correspond to detected local objects. The global node captures holistic scene context and serves as a complementary cue when local detection is incomplete or unstable. The local nodes, on the other hand, provide object-level semantic information that supports interpretable and fine-grained correspondence modeling. The combination of global and local nodes therefore enables the representation to balance robustness and discrimination in cross-view matching.

For Faster R-CNN, the global feature $\bm{r}^{(0)}$ is derived by fusing the first four convolutional outputs of the backbone (excluding the final pooling layer). Bounding box features are extracted after ROI pooling and pyramid pooling, before the final classification fully connected layer. The resulting feature dimension is \((1,1024)\), i.e., \( D=1024 \).  

For YOLOv8, since no direct interface is provided to output the global feature map, this study follows the authors’ recommendation~\cite{ref47} and employs a hook function to extract features from the two layers of the 20th Concat structure, which are fused to form the global feature. For bounding box features, this study modified the non-maximum suppression function in opts.py of the Ultralytics package to additionally return boxes\_features. These features have size \((1,64)\), i.e., \( D=64 \), which partly explains YOLO’s faster computation speed.  

\subsection{Object-aware dual-graph construction}

It is well established that target regions in an image exhibit both semantic and spatial relationships. To capture these dependencies, this study constructs two types of graphs: the spatial graph \( \bm{G}_{\mathrm{sp}}=(\bm{V}_{\mathrm{sp}},\bm{W}_{\mathrm{sp}}) \) and the semantic graph \( \bm{G}_{\mathrm{se}}=(\bm{V}_{\mathrm{se}},\bm{W}_{\mathrm{se}}) \), where \( \bm{W}_{\mathrm{sp}} \) and \( \bm{W}_{\mathrm{se}} \) are weighted adjacency matrices. The spatial graph encodes positional relations among target regions, derived directly from their bounding boxes. In contrast, the semantic graph models higher-level associations beyond spatial proximity, incorporating attributes such as category labels, confidence scores, and inter-node interactions, which are implicitly learned during training.

\subsubsection{Spatial graph}
The spatial graph \( \bm{G}_{\mathrm{sp}} \) represents positional relationships among target regions within an image. It introduces geometric constraints by encoding the relative layout of detected objects, which is useful when the spatial arrangement of scene elements is relatively stable. However, geometric relations alone may become unreliable under cross-view and cross-modal variations, where perspective differences can distort object layouts. Its nodes \( \bm{V}_{\mathrm{sp}} \) are derived from the regional feature set \( \bm{V} \), while the weighted adjacency matrix \( \bm{W}_{\mathrm{sp}} \) encodes pairwise spatial dependencies. Since target regions vary in size, exact distances are difficult to compute. This study approximates spatial relations using the overlap area, centroid distance, and aspect ratios of bounding boxes.~\cite{ref52} Specifically, this study adopts the Complete Intersection over Union (CIoU), which can be computed directly from bounding box parameters and jointly reflects overlap, centroid distance, and aspect ratio consistency between regions.

To further enhance robustness, cosine similarity is incorporated, as it is insensitive to vector magnitude changes caused by lighting, occlusion, or scale variations, focusing instead on directional similarity. The contribution of cosine similarity in this formulation is further examined in the ablation study in Section~\ref{sec:ablation_cosine}. Thus, the weight \(W_{\mathrm{sp}}(i,j) \) between regions \(i\) and \(j\) is defined as
\begin{equation}
W_{\mathrm{sp}}(i,j)=
\begin{cases}
\text{CIoU}(i,j)\cdot \cos(\bm{v}_i,\bm{v}_j) & \text{CIoU}(i,j)\geq \theta\\
0 & \text{otherwise}
\end{cases}
\label{eq:weight}
\end{equation} where \( \cos(\cdot) \) denotes cosine similarity, and \( \theta \) is a predefined threshold.

\subsubsection{Semantic graph}

\begin{figure}[t]
    \centering
    \includegraphics[width=0.9\linewidth]{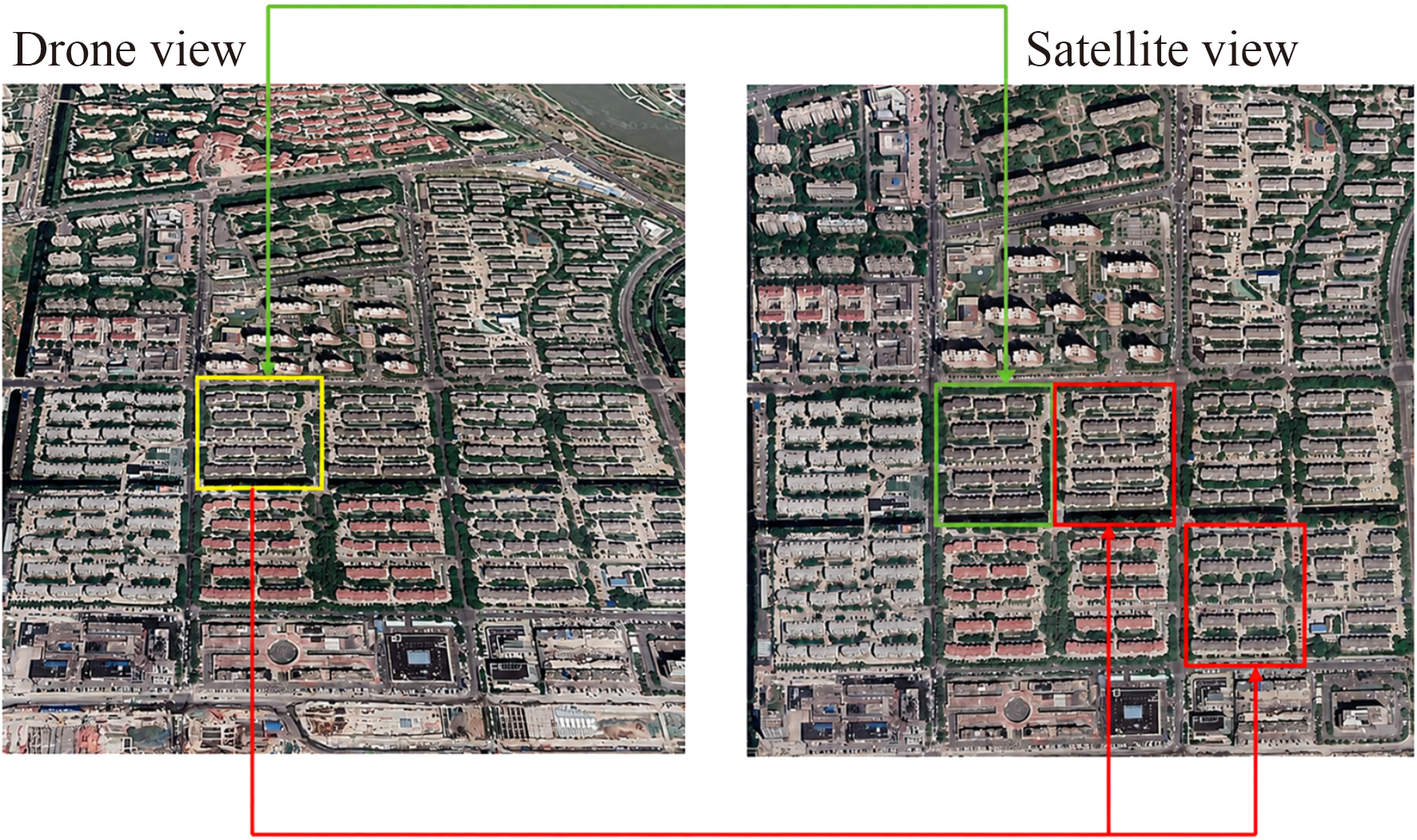}
    \caption{Illustration of matching ambiguities in drone-to-satellite image retrieval.}
    \label{fig4}
\end{figure}

The semantic graph \( \bm{G}_{\mathrm{se}} \) captures higher-level associations among target regions and is introduced to alleviate the instability of geometric relations under cross-view transformations, where spatial layouts may vary significantly between drone and satellite images. Compared with the spatial graph, the semantic graph focuses more on the intrinsic properties and relationships of objects rather than their absolute locations, thereby providing complementary semantic cues for cross-modal matching. Beyond spatial proximity, targets may share semantic similarities, such as recurring structural patterns, which are often preserved across drone and satellite views. However, repetitive semantic patterns may also introduce ambiguities and lead to incorrect matches, as illustrated in Fig.~\ref{fig4}, where the target region is highlighted with a yellow bounding box, the correct match is indicated in green, and the incorrect match is marked in red. Despite this challenge, semantic relations remain generally more robust than purely geometric configurations when spatial layouts differ significantly between modalities. Therefore, this study constructs semantic graphs for individual images to enable the network to learn more reliable cross-modal semantic correspondences between drone and satellite imagery.

In the semantic graph \( \bm{G}_{\mathrm{se}} \), the node set
\( \bm{V}_{\mathrm{se}} \) is constructed from the regional feature set
\( \bm{V} \), and the weighted adjacency matrix
\( \bm{W}_{\mathrm{se}} \) encodes the connection strength between regions.
Inspired by Ref.~\citenum{ref57},  this study defines the \((i,j)\)th element of \( \bm{W}_{\mathrm{se}} \) as

\begin{equation}
W_{\mathrm{se}}(i,j)
=
\bm{\psi}(\bm{v}_{\mathit{i}})^{\mathrm{T}}
\bm{\phi}(\bm{v}_{\mathit{j}})
\label{eq:w_se}
\end{equation}
where
\( \bm{\psi}(\bm{v}_{\mathit{i}})=\bm{W}_{\psi}\bm{v}_{\mathit{i}} \)
and
\( \bm{\phi}(\bm{v}_{\mathit{j}})=\bm{W}_{\phi}\bm{v}_{\mathit{j}} \)
are two feature embeddings, and
\( \bm{W}_{\psi} \) and \( \bm{W}_{\phi} \)
are learnable projection matrices.
The semantic graph \( \bm{G}_{\mathrm{se}} \) is fully connected, and each
edge weight reflects the potential semantic relationship between target
regions.

Thus, this study obtains the drone graph \( \bm{G}_{\mathrm{u}}=(\bm{G}_{\mathrm{u}}^{\mathrm{sp}},\bm{G}_{\mathrm{u}}^{\mathrm{se}}) \) and the satellite graph \( \bm{G}_{\mathrm{s}}=(\bm{G}_{\mathrm{s}}^{\mathrm{sp}},\bm{G}_{\mathrm{s}}^{\mathrm{se}}) \).

\subsubsection{Design rationale and distinction from prior graph formulations}

The proposed dual-graph design is motivated by the complementary nature of geometric and semantic cues in cross-domain remote sensing localization. A single graph is often insufficient to capture both aspects effectively. The spatial graph provides geometric constraints by modeling object layout, while the semantic graph compensates for the instability of geometric relations under cross-view and cross-modal variations. In this way, the two graphs jointly improve robustness and discriminative capability.

In addition, the graph nodes are constructed from detected objects rather than dense patch tokens, since detector-derived regions are more closely aligned with stable, localization-relevant targets and provide a more compact and interpretable representation. The edge formulations are also task-oriented: the spatial graph uses CIoU multiplied by cosine similarity to combine geometric structure with feature consistency, while the semantic graph adopts a learnable affinity to model higher-level semantic relations between object regions. Unlike prior graph formulations in image-text matching, which mainly focus on visual-linguistic alignment, the proposed method targets drone-to-satellite localization by building object-aware structural representations within each image. Rather than introducing a fundamentally new generic graph operator, the proposed method focuses on an object-aware dual-graph representation tailored to cross-domain remote sensing localization.

The contribution of the proposed method is therefore not a fundamentally new generic graph operator, but a task-specific object-aware dual-graph representation tailored to cross-domain remote sensing geo-localization.

\subsection{Reasoning and embedding with graph neural networks}

To perform relational reasoning, this study employs TransformerConv as the final graph neural operator in OGMN, so that the theoretical formulation is consistent with the implementation described in Section~\ref{V-C}. Given a graph \( \bm{G}=(\bm{V},\bm{W}) \) with \(K\) nodes, where 
\( \bm{V}=[\bm{v}_1;\bm{v}_2;\dots;\bm{v}_K]\in\mathbb{R}^{K\times D} \) denotes the stacked node feature matrix and 
\( \bm{W}\in\mathbb{R}^{K\times K} \) denotes the weighted adjacency matrix, the neighborhood of node \(i\) is defined as 
\( \mathcal{N}(i)=\{j\mid W_{ij}\neq 0\}\cup\{i\} \). 
For the \(h\)th attention head, the query, key, and value projections are computed as

\begin{equation}
\bm{q}_i^{h}=\bm{W}_{\mathrm{Q}}^{h}\bm{v}_i,\quad
\bm{k}_j^{h}=\bm{W}_{\mathrm{K}}^{h}\bm{v}_j,\quad
\bm{z}_j^{h}=\bm{W}_{\mathrm{V}}^{h}\bm{v}_j 
\end{equation} where \( \bm{W}_{\mathrm{Q}}^{h} \), \( \bm{W}_{\mathrm{K}}^{h} \), and \( \bm{W}_{\mathrm{V}}^{h} \) are learnable projection matrices.

The attention coefficient between node \(i\) and its neighbor \(j\) is calculated as

\begin{equation}
e_{ij}^{h}=
\frac{(\bm{q}_i^{h})^{\mathrm{T}}\bm{k}_j^{h}}{\sqrt{d_h}}+\eta(W_{ij})
\end{equation}

\begin{equation}
\alpha_{ij}^{h}=
\operatorname{softmax}_{j\in\mathcal{N}(i)}(e_{ij}^{h})
\end{equation} where \(d_h\) is the feature dimension of each attention head, and 
\( \eta(W_{ij}) \) denotes the edge-weight bias derived from the spatial or semantic adjacency matrix.

The updated node representation is then obtained by multi-head aggregation
\begin{equation}
\bm{v}_i^{*}
=
\sigma\!\left(
\bm{W}_{\mathrm{O}}
\operatorname{Concat}_{h=1}^{H}
\sum_{j\in\mathcal{N}(i)}
\alpha_{ij}^{h}\bm{z}_j^{h}
\right)
+\bm{v}_i 
\label{eq:gnn_layer}
\end{equation} where \(H=4\) in the first TransformerConv layer of the implementation, \( \bm{W}_{\mathrm{O}} \) is a learnable output projection matrix, the residual connection is used to preserve the original node representation and \(\sigma(\cdot)\) represents the Sigmoid function. This formulation describes the TransformerConv operator adopted in the final model, while Graph Convolutional Networks (GCN), Graph Attention Networks (GAT), and GraphSAGE are only used as comparative variants in the ablation study.

\subsubsection{Reasoning and embedding for drone graph}
The drone spatial and semantic graphs are separately processed by GNNs and then fused to form the drone graph representation
\begin{equation}
\bm{V}_{\mathrm{u}}^{\mathrm{sp}*}
=
\operatorname{GNN}_{\mathrm{sp}}
(\bm{V}_{\mathrm{u}}^{\mathrm{sp}}, \bm{W}_{\mathrm{u}}^{\mathrm{sp}})
\end{equation}

\begin{equation}
\bm{V}_{\mathrm{u}}^{\mathrm{se}*}
=
\operatorname{GNN}_{\mathrm{se}}
(\bm{V}_{\mathrm{u}}^{\mathrm{se}}, \bm{W}_{\mathrm{u}}^{\mathrm{se}})
\end{equation}

\begin{equation}
\bm{V}_{\mathrm{u}}^{*}
=
\frac{
\bm{V}_{\mathrm{u}}^{\mathrm{sp}*}
+
\bm{V}_{\mathrm{u}}^{\mathrm{se}*}
}{2}
\end{equation}

Compared with conventional global feature learning, graph-based relational reasoning over detected object regions has received relatively limited attention in cross-view image matching and localization. Considering computational efficiency, this study evaluates several simple but effective GNN variants, including GCN,~\cite{ref54} GAT,~\cite{ref55} GraphSAGE,~\cite{ref56} and TransformerConv.~\cite{ref57} Among them, TransformerConv achieves the best performance in the localization task.

For node embedding aggregation, this study further compares Gated Recurrent Units (GRU),~\cite{ref58} Graph Embedding Module (GEM),~\cite{ref60} and mean pooling. Inspired by Ref.~\citenum{ref59}, GEM proves most effective for localization. Thus, the final drone embedding is obtained as
\begin{equation}
\bm{I}_{\mathrm{u}} = \operatorname{GEM}(\bm{V}_{\mathrm{u}}^{*})
\end{equation}

\subsubsection{Reasoning and embedding for satellite graph}
The satellite graph follows the same reasoning pipeline as the drone graph. Its spatial and semantic branches are separately processed by GNNs, then fused and aggregated to obtain the final satellite embedding

\begin{equation}
\bm{V}_{\mathrm{s}}^{\mathrm{sp}*}
=
\operatorname{GNN}_{\mathrm{sp}}
(\bm{V}_{\mathrm{s}}^{\mathrm{sp}}, \bm{W}_{\mathrm{s}}^{\mathrm{sp}})
\end{equation}

\begin{equation}
\bm{V}_{\mathrm{s}}^{\mathrm{se}*}
=
\operatorname{GNN}_{\mathrm{se}}
(\bm{V}_{\mathrm{s}}^{\mathrm{se}}, \bm{W}_{\mathrm{s}}^{\mathrm{se}})
\end{equation}

\begin{equation}
\bm{V}_{\mathrm{s}}^{*}
=
\frac{
\bm{V}_{\mathrm{s}}^{\mathrm{sp}*}
+
\bm{V}_{\mathrm{s}}^{\mathrm{se}*}
}{2}
\end{equation}

\begin{equation}
\bm{I}_{\mathrm{s}}
=
\operatorname{GEM}(\bm{V}_{\mathrm{s}}^{*})
\end{equation}

\subsection{Matching of drone node graphs and satellite node graphs}

After acquiring drone and satellite images, the next step is to perform cross-view matching. Prior work such as SimGNN~\cite{ref45} combined siamese networks with graph convolution, achieving graph matching via node-level embeddings and attention mechanisms. Ref.~\citenum{ref62} further highlighted the importance of node matching in graph similarity computation. Motivated by these studies, this study designs a customized graph node matching method tailored to aerial image object detection. This approach enables the model to learn region correspondences and relationships across spatiotemporal domains and viewpoints in heterogeneous images.

To ensure accuracy while maintaining efficiency during inference, this study employs a joint loss function-integrating node matching loss, embedding loss, and classification loss-only during training, and omits it at test time.

\subsubsection{Graph node matching loss}

To align salient regions across cross-source multi-view images, this study proposes a multi-granularity graph node matching loss driven by global features. It introduces primary nodes (via global aggregation) for coarse alignment and secondary nodes for fine-grained matching. Different from conventional pairwise contrastive learning, which usually applies a uniform matching objective to all candidate pairs, the proposed loss explicitly separates global-level alignment from object-level correspondence. The primary node branch provides a coarse matching signal for overall cross-view consistency, while the secondary node branch focuses on fine-grained alignment between local object regions. In addition, valid secondary pairs are not selected arbitrarily, but are constrained by class consistency and coordinate proximity, and further weighted by detection confidence. This design makes the matching objective better suited to the structural characteristics of cross-domain remote sensing localization.

\textbf{Primary Node Matching Loss.}
Primary node features are generated through global aggregation. The loss is defined as
\begin{equation}
L_{\mathrm{main}} =
-w_{\mathrm{global}}
\left[
y\ln \sigma(s_{\mathrm{global}})
+
(1-y)\ln(1-\sigma(s_{\mathrm{global}}))
\right]
\label{eq:loss_main}
\end{equation} where \(s_{\mathrm{global}}\) denotes the temperature-scaled cosine similarity between the drone and satellite primary nodes, \(y\in\{0,1\}\) is the binary matching label. The logarithm \(\ln(\cdot)\) denotes the natural logarithm. The similarity score and confidence-based weighting factor are computed as

\begin{equation}
s_{\mathrm{global}}
=
\frac{
(\bm{v}_{\mathrm{dro}}^{*})^{\mathrm{T}}
\bm{v}_{\mathrm{sat}}^{*}
}
{
\tau
\|\bm{v}_{\mathrm{dro}}^{*}\|
\|\bm{v}_{\mathrm{sat}}^{*}\|
},
\quad
\tau\in\mathbb{R}^{+}
\end{equation}

\begin{equation}
w_{\mathrm{global}}
=
\sqrt{
c_{\mathrm{dro}}\cdot c_{\mathrm{sat}}
}
\end{equation} where \(\tau\) is the temperature parameter that controls the smoothness of the similarity distribution. A smaller \(\tau\) produces a sharper similarity response, while a larger \(\tau\) results in a smoother distribution. \(w_{\mathrm{global}}\) denotes the confidence-based weighting factor for the primary node pair, which measures the reliability of the matching prediction by combining the detection confidence scores of the drone and satellite primary nodes. Specifically, \(c_{\mathrm{dro}}\) and \(c_{\mathrm{sat}}\) represent the detection confidence scores of the primary nodes in drone and satellite images, respectively.

\textbf{Secondary Node Matching Loss.}
Secondary nodes are all non-primary targets. Let
$\bm{p}_{i}^{\mathrm{dro}}\in[0,1]^2$ and
$\bm{p}_{j}^{\mathrm{sat}}\in[0,1]^2$
denote the normalized center coordinates of the $i$th drone box and the
$j$th satellite box, respectively, normalized by the image width and height.
A pair $(i,j)$ is considered a valid candidate pair if

\begin{equation}
\left\|
\bm{p}_{i}^{\mathrm{dro}}
-
\bm{p}_{j}^{\mathrm{sat}}
\right\|_2
\leq
\epsilon_{\mathrm{coord}}
\end{equation}
\begin{equation}
\mathrm{cls}_{i}^{\mathrm{dro}}
=
\mathrm{cls}_{j}^{\mathrm{sat}}
\end{equation} where $\epsilon_{\mathrm{coord}}$ is the coordinate distance threshold used
to filter potential secondary node pairs based on spatial consistency, and
$\mathrm{cls}_{i}^{\mathrm{dro}}$ and
$\mathrm{cls}_{j}^{\mathrm{sat}}$ denote the predicted class labels of the
$i$th drone box and the $j$th satellite box, respectively. The class
consistency constraint ensures that only secondary nodes belonging to the
same predicted category are considered as valid candidate pairs.

The coordinate discrepancy between a candidate pair $(i,j)$ is defined as

\begin{equation}
D_{\mathrm{coord}}^{(i,j)}
=
\left\|
\bm{p}_{i}^{\mathrm{dro}}
-
\bm{p}_{j}^{\mathrm{sat}}
\right\|_2 
\end{equation}

The confidence weight for a valid pair $(i,j)$ is defined as

\begin{equation}
W_{\mathrm{sub}}^{(i,j)}
=
\min
\left(
c_{i}^{\mathrm{dro}},
c_{j}^{\mathrm{sat}}
\right)
\cdot
\exp
\left(
-\beta D_{\mathrm{coord}}^{(i,j)}
\right)
\end{equation} where $c_{i}^{\mathrm{dro}}$ and $c_{j}^{\mathrm{sat}}$ are the detection
confidences of the corresponding secondary nodes. The parameter $\beta$
controls the decay rate of the coordinate discrepancy penalty, where a
larger $\beta$ assigns lower weights to node pairs with larger spatial
deviations.

The pairwise similarity score is defined as

\begin{equation}
s_{ij}
=
\frac{
\cos
\left(
\bm{v}_{i}^{\mathrm{dro}},
\bm{v}_{j}^{\mathrm{sat}}
\right)
}
{\tau_{\mathrm{sub}}},
\qquad
\tau_{\mathrm{sub}}\in\mathbb{R}^{+}
\end{equation} where $\bm{v}_{i}^{\mathrm{dro}}$ and
$\bm{v}_{j}^{\mathrm{sat}}$ denote the feature embeddings of the
corresponding secondary nodes, and $\tau_{\mathrm{sub}}$ is the temperature
parameter controlling the scale of the similarity distribution.

The secondary node matching loss is formulated as a confidence-weighted
binary cross-entropy loss:

\begin{equation}
\begin{aligned}
L_{\mathrm{sub}}
=
-\frac{1}
{\sum_{i,j}W_{\mathrm{sub}}^{(i,j)}}
\sum_{i,j}
W_{\mathrm{sub}}^{(i,j)}
\Big[
&y_{ij}\ln\sigma(s_{ij})
\\
&+
(1-y_{ij})
\ln\left(1-\sigma(s_{ij})\right)
\Big]
\end{aligned}
\label{eq:loss_sub}
\end{equation} where $y_{ij}\in\{0,1\}$ indicates whether the pair $(i,j)$ represents a
positive correspondence, determined by the pseudo-labeling strategy
described in Section~\ref{V-C}. $L_{\mathrm{sub}}$ denotes the secondary node
matching loss, which optimizes the correspondence prediction between
non-primary targets by aggregating confidence-weighted pairwise matching
errors.

\subsubsection{Graph embedding matching loss}

Effective mining of positive and negative pairs is crucial for discriminative feature learning.~\cite{ref59} This study adopts Circle Loss \cite{ref63} and further proposes a global-local coupling strategy to leverage both contextual and object-level features. This loss serves as the main objective for cross-view retrieval, as the final localization decision is based on graph-level embedding similarity rather than individual node correspondence alone. In contrast, the node matching loss provides auxiliary supervision to improve fine-grained alignment between salient regions. The proposed global-local coupling strategy avoids over-reliance on either holistic scene context or isolated object cues, enabling the learned embedding to simultaneously preserve global consistency and local discriminative information.

The global similarity is defined as

\begin{equation}
S_{\mathrm{g}}
=
\cos
\left(
\bm{v}_{\mathrm{global}}^{\mathrm{dro}},
\bm{v}_{\mathrm{global}}^{\mathrm{sat}}
\right)
\end{equation} where \( \bm{v}_{\mathrm{global}}^{\mathrm{dro}} \) and \( \bm{v}_{\mathrm{global}}^{\mathrm{sat}} \) denote the global graph embeddings of the drone and satellite images, respectively.

The local similarity is defined as

\begin{equation}
S_{\mathrm{l}}
=
\frac{1}{K_{\mathrm{u}}}
\sum_{k=1}^{K_{\mathrm{u}}}
\max_{1\leq j\leq K_{\mathrm{s}}}
\left(
\cos
(
\bm{v}_{\mathrm{u},k}^{*},
\bm{v}_{\mathrm{s},j}^{*}
)
\right)
\end{equation} where \(K_{\mathrm{u}}\) and \(K_{\mathrm{s}}\) denote the numbers of nodes in the drone and satellite graphs, respectively, and \( \bm{v}_{\mathrm{u},k}^{*} \) and \( \bm{v}_{\mathrm{s},j}^{*} \) represent the GNN-updated node embeddings. The local similarity measures the correspondence between object-level features by searching the most similar satellite node for each drone node.

To incorporate the object-level overlap and spatial consistency between cross-view image pairs, an overlap-aware weighting strategy is introduced

\begin{equation}
w_{\mathrm{o}}
=
\frac{T_{\mathrm{common}}}
{T_{\mathrm{UAV}}+T_{\mathrm{Sat}}}
\cdot
\left(
1+
\frac{
\operatorname{IoU}_{\mathrm{global}}
}
{
1+\operatorname{IoU}_{\mathrm{global}}
}
\right)
\end{equation} where \(T_{\mathrm{common}}\) denotes the number of shared objects between the drone and satellite graphs, while \(T_{\mathrm{UAV}}\) and \(T_{\mathrm{Sat}}\) represent the total numbers of detected objects in the drone and satellite images, respectively. \( \operatorname{IoU}_{\mathrm{global}} \) denotes the Intersection-over-Union (IoU) between the global regions of the drone and satellite views, which measures the degree of spatial overlap between the two images. The weighting factor \(w_{\mathrm{o}}\) adaptively emphasizes image pairs with stronger object-level overlap and higher spatial consistency.

The final graph embedding matching loss is formulated as

\begin{equation}
L_{\mathrm{M}}
=
\begin{cases}
\ln
\left[
1+\exp
\left(
-\gamma w_{\mathrm{o}}
(S_{\mathrm{g}}+S_{\mathrm{l}}-m_{\mathrm{p}})
\right)
\right],
& y=1,
\\[6pt]
\ln
\left[
1+\exp
\left(
\gamma
(S_{\mathrm{g}}+S_{\mathrm{l}}-m_{\mathrm{n}})
\right)
\right],
& y=0.
\end{cases}
\end{equation} where \(y\in\{0,1\}\) indicates whether the drone-satellite image pair is a positive or negative sample. The parameters \(m_{\mathrm{p}}\) and \(m_{\mathrm{n}}\) represent the positive and negative margins, respectively, and are empirically set to \(m_{\mathrm{p}}=1.5\) and \(m_{\mathrm{n}}=-0.5\). The parameter \(\gamma\) is the scale factor controlling the optimization strength of the Circle Loss objective, where a larger value increases the penalty for samples that violate the desired similarity margins.

\subsubsection{Graph classification loss}

To enhance scene-level discrimination, this study employs an attention-based graph classification loss.~\cite{ref64} This loss acts as a scene-level regularization term rather than the primary retrieval objective, encouraging the learned graph representation to preserve category-level discriminative information.

Given a graph \( \bm{G}=(\bm{V},\bm{W}) \) with node features \( \bm{V}\in\mathbb{R}^{K\times D} \), this study aggregates a global representation Multi-Head Attention (MHA)~\cite{ref14}

\begin{equation}
\operatorname{Concat}
\left(
\bm{\mathrm{head}}_{1},
\bm{\mathrm{head}}_{2},
\dots,
\bm{\mathrm{head}}_{H}
\right)
\bm{W}_{\mathrm{O}}
\end{equation}

For the \(i\)th attention head, the attention output is computed as

\begin{equation}
\bm{\mathrm{head}}_{i}
=
\operatorname{softmax}
\left(
\frac{
\bm{Q}_{i}\bm{K}_{i}^{\mathrm{T}}
}
{\sqrt{d_{\mathrm{k}}}}
\right)
\bm{V}_{i}
\end{equation} where 
\[
\bm{Q}_{i}=\bm{V}\bm{W}_{\mathrm{Q}}^{i},
\quad
\bm{K}_{i}=\bm{V}\bm{W}_{\mathrm{K}}^{i},
\quad
\bm{V}_{i}=\bm{V}\bm{W}_{\mathrm{V}}^{i}
\]
are the query, key, and value matrices of head \(i\), respectively, and \(d_{\mathrm{k}}\) is the key dimension.

The global feature is classified as

\begin{equation}
\bm{p}
=
\operatorname{softmax}
(
\bm{W}_{C}
\bm{h}_{\mathrm{global}}
+
\bm{b}_{C}
)
\end{equation} where \( \bm{p}\in\mathbb{R}^{C} \) is the predicted probability distribution over \(C\) classes.

This study adopts cross-entropy with label smoothing

\begin{equation}
\tilde{y}_{c}
=
(1-\alpha)y_c+\frac{\alpha}{C},
\qquad c=1,2,\dots,C
\end{equation}

\begin{equation}
L_{\mathrm{cls}}
=
-\sum_{c=1}^{C}
\tilde{y}_{c}\ln p_c
\end{equation} where \(\alpha=0.1\) is the smoothing factor, \(y_c\) is the one-hot ground-truth label, and \(p_c\) is the predicted probability for class \(c\).

\subsubsection{Multi-task joint optimization}

Inspired by Ref.~\citenum{ref66}, this study adopts a dynamic weighting strategy to balance the graph node matching loss \(L_{\mathrm{node}}\), graph embedding matching loss \(L_{\mathrm{M}}\), and graph classification loss \(L_{\mathrm{cls}}\) during training

\begin{equation}
L_{\mathrm{total}}
=
\lambda_1L_{\mathrm{node}}
+
\lambda_2L_{\mathrm{M}}
+
\lambda_3L_{\mathrm{cls}}
\label{eq:total_loss_final}
\end{equation} where \( \lambda_1, \lambda_2, \lambda_3 \) are task weights for balancing the three objectives. This strategy is used to stabilize optimization and coordinate the contributions of different losses, rather than constituting the primary methodological novelty of this work.

\textbf{Weight Initialization.}
At the beginning of training, relatively larger weight is assigned to node matching in order to establish more stable object-level representations. The weights are initialized as
\begin{equation}
\lambda_1 : \lambda_2 : \lambda_3 = 3 : 2 : 2
\end{equation}

\textbf{Dynamic Weight Adjustment.}
During training, the weights are adjusted every $T$ epochs according to recall performance on the validation set
\begin{equation}
\lambda_{i}^{\prime} = \lambda_{i} \cdot (1 + \Delta R_{i})
\label{eq:lambda_update}
\end{equation} where $\Delta R_i$ denotes the recall gain associated with task $i$.

\textbf{Weight Normalization.}
After each update, the task weights are normalized to avoid imbalance
\begin{equation}
\lambda_1 + \lambda_2 + \lambda_3 = 1
\end{equation}
This helps prevent any single task from dominating the optimization process.

\subsection{Algorithm overview}
The proposed Algorithm~\ref{alg:alg1} is designed for cross-domain remote sensing image localization. The pipeline begins with object detection to extract regional features from drone and satellite images. These features are then represented as graph nodes, with spatial and semantic relations encoded as weighted adjacency matrices. Separate GNNs process the spatial and semantic graphs, and the results are aggregated using a Graph Embedding Module (GEM) to generate compact embeddings. Localization is achieved by computing the similarity between drone and satellite embeddings. To ensure robustness, Quality-Aware Template Matching (QATM)~\cite{ref16} is employed as a fallback when no valid targets are detected. The QATM fallback is triggered only when no valid target regions are detected in either the drone or satellite image. This setting reflects that the proposed graph-based pipeline depends on the availability of detector-derived object regions. Therefore, QATM is introduced as a robustness-oriented backup strategy for rare failure cases, rather than as part of the main inference path of OGMN.

\begin{algorithm}[t]
\caption{Object-aware Graph Matching Network (OGMN).}
\label{alg:alg1}

\algsetup{linenodelimiter=.}

\begin{algorithmic}[1]

\renewcommand{\algorithmicrequire}{\textbf{Input:}}
\renewcommand{\algorithmicensure}{\textbf{Output:}}

\REQUIRE Drone images $\mathcal{U}$; Satellite images $\mathcal{S}$
\ENSURE Drone and Satellite graph embeddings $\bm{I}_{\mathrm{u}}, \bm{I}_{\mathrm{s}}$

\STATE {// \textbf{Stage 1}: Feature extraction \& region proposal}
\STATE Extract region features $\bm{R}_{\mathrm{u}}$ and $\bm{R}_{\mathrm{s}}$ using an object detector

\IF{$\bm{R}_{\mathrm{u}}=\varnothing$ \OR $\bm{R}_{\mathrm{s}}=\varnothing$}
    \STATE Execute QATM-based template matching as fallback
\ELSE

\STATE {// \textbf{Stage 2}: Node construction (per view)}
\STATE Build node features $\bm{V}_{\mathrm{u}}$ from $\bm{R}_{\mathrm{u}}$ and $\bm{V}_{\mathrm{s}}$ from $\bm{R}_{\mathrm{s}}$ via
$\bm{v}^{(i)}
\leftarrow
\bm{W}_{\mathrm{f}}\bm{r}^{(i)}
+\bm{b}_{\mathrm{f}}$

\STATE {// \textbf{Stage 3}: Edge construction (per view)}
\STATE Compute $\bm{W}_{\mathrm{u}}^{\mathrm{sp}}$ from drone boxes and $\bm{W}_{\mathrm{s}}^{\mathrm{sp}}$ from satellite boxes using Eq.~\eqref{eq:weight}

\STATE Compute $\bm{W}_{\mathrm{u}}^{\mathrm{se}}$ from drone nodes and $\bm{W}_{\mathrm{s}}^{\mathrm{se}}$ from satellite nodes using Eq.~\eqref{eq:w_se}

\STATE {// \textbf{Stage 4}: Multi-graph reasoning and aggregation}

\STATE $\bm{V}_{\mathrm{u}}^{\mathrm{sp}*}
\leftarrow
\operatorname{GNN}_{\mathrm{sp}}
(\bm{V}_{\mathrm{u}},\bm{W}_{\mathrm{u}}^{\mathrm{sp}})$

\STATE $\bm{V}_{\mathrm{u}}^{\mathrm{se}*}
\leftarrow
\operatorname{GNN}_{\mathrm{se}}
(\bm{V}_{\mathrm{u}},\bm{W}_{\mathrm{u}}^{\mathrm{se}})$

\STATE $\bm{I}_{\mathrm{u}}
\leftarrow
\operatorname{GEM}
\left(
\frac{1}{2}
(
\bm{V}_{\mathrm{u}}^{\mathrm{sp}*}
+
\bm{V}_{\mathrm{u}}^{\mathrm{se}*}
)
\right)$

\STATE $\bm{V}_{\mathrm{s}}^{\mathrm{sp}*}
\leftarrow
\operatorname{GNN}_{\mathrm{sp}}
(\bm{V}_{\mathrm{s}},\bm{W}_{\mathrm{s}}^{\mathrm{sp}})$

\STATE $\bm{V}_{\mathrm{s}}^{\mathrm{se}*}
\leftarrow
\operatorname{GNN}_{\mathrm{se}}
(\bm{V}_{\mathrm{s}},\bm{W}_{\mathrm{s}}^{\mathrm{se}})$

\STATE $\bm{I}_{\mathrm{s}}
\leftarrow
\operatorname{GEM}
\left(
\frac{1}{2}
(
\bm{V}_{\mathrm{s}}^{\mathrm{sp}*}
+
\bm{V}_{\mathrm{s}}^{\mathrm{se}*}
)
\right)$

\ENDIF

\STATE Rank candidates based on $\operatorname{sim}(\bm{I}_{\mathrm{u}},\bm{I}_{\mathrm{s}})$

\end{algorithmic}
\end{algorithm}

\subsection{Discussion on robustness and failure modes}

Although the proposed object-aware dual-graph formulation improves structural modeling for cross-domain remote sensing localization, its performance still depends on the quality of object detection. When salient targets are missed by the detector, the resulting graph may contain insufficient nodes or incomplete structural information, which weakens subsequent spatial and semantic reasoning. Conversely, false detections may introduce noisy or irrelevant nodes, thereby disturbing graph construction and degrading the reliability of inter-region relations.

To alleviate these issues, the framework incorporates a global node to preserve holistic scene context even when local object detection is incomplete. In addition, a QATM-based fallback branch is introduced for rare cases where no valid targets are detected. These mechanisms improve robustness to detector failure to some extent. However, they cannot fully eliminate the effect of error propagation from the detection stage to graph reasoning and final retrieval. Therefore, the overall robustness of the method remains influenced by detector precision and recall, especially under severe appearance change, occlusion, or extreme cross-view discrepancy.

\section{Experiments}
To rigorously evaluate the effectiveness of the proposed model and quantify the contribution of each module, this study adopts widely used retrieval-based metrics,~\cite{ref19,ref20} including Recall@$K$, Recall@$1$, and Average Precision (AP). These indicators jointly provide a comprehensive assessment of retrieval accuracy, robustness, and generalization.  

\begin{table*}[t]
  \captionsetup{width=\linewidth}
  \centering
  \caption{Official dataset splits and basic statistics of the datasets used in this study.}
  \label{tab:table5}

  \scriptsize
  \setlength{\tabcolsep}{4pt}
  \renewcommand{\arraystretch}{1.15}

  \begin{tabularx}{\linewidth}
    {l *{4}{>{\centering\arraybackslash}X}}
    \toprule

    Datasets
    & University-1652
    & SUES-200
    & DenseUAV
    & IRVL328 \\

    \midrule

    Training
    & $701 \times 54 + 701 \times 1$
    & $480 \times 50 + 480 \times 1$
    & $2256 \times 3 + 2256 \times 6$
    & $200 \times 50 + 200 \times 1$ \\

    Testing
    & $701 \times 54 + 701 \times 1$
    & $320 \times 50 + 320 \times 1$
    & $2256 \times 3 + 3033 \times 6$
    & $128 \times 50 + 128 \times 1$ \\

    Platform
    & Drone, Ground, Satellite
    & Drone, Satellite
    & Drone, Satellite
    & Drone, Satellite \\

    GeoTag
    & $\checkmark$
    & $\checkmark$
    & $\checkmark$
    & $\checkmark$ \\

    Evaluation
    & Recall@$K$ \& AP
    & Recall@$K$ \& AP
    & Recall@$K$ \& AP
    & Recall@$K$ \& AP \\

    \bottomrule
  \end{tabularx}

\end{table*}

\subsection{Datasets}
A summary of the datasets employed in this study is provided in Table~\ref{tab:table5}.  

\subsubsection{University-1652}
A large-scale multi-view cross-source benchmark containing satellite, drone, and
ground-view images of 1652 buildings from 72 universities worldwide. Drone-view
data are generated from oblique projections, enabling both
Drone$\rightarrow$Satellite localization and
Satellite\allowbreak$\rightarrow$Drone navigation tasks.

\subsubsection{SUES-200}
A real-world multi-altitude dataset collected under diverse natural conditions. Drone images were captured at 150, 200, 250, and 300 m altitudes. It incorporates variations in illumination and shadows, making it a challenging benchmark for testing the robustness of localization and navigation algorithms.  

\subsubsection{DenseUAV}
A multi-altitude and multi-temporal dataset collected from 14 universities in Zhejiang Province. It contains drone-view images at 80, 90, and 100 m, paired with satellite imagery from both 2020 and 2022. Owing to substantial spatiotemporal variations, this dataset is mainly used to evaluate Drone$\rightarrow$Satellite self-localization.  

\subsubsection{IRVL328}
\begin{figure}[t]
    \centering
    \includegraphics[width=0.75\linewidth]{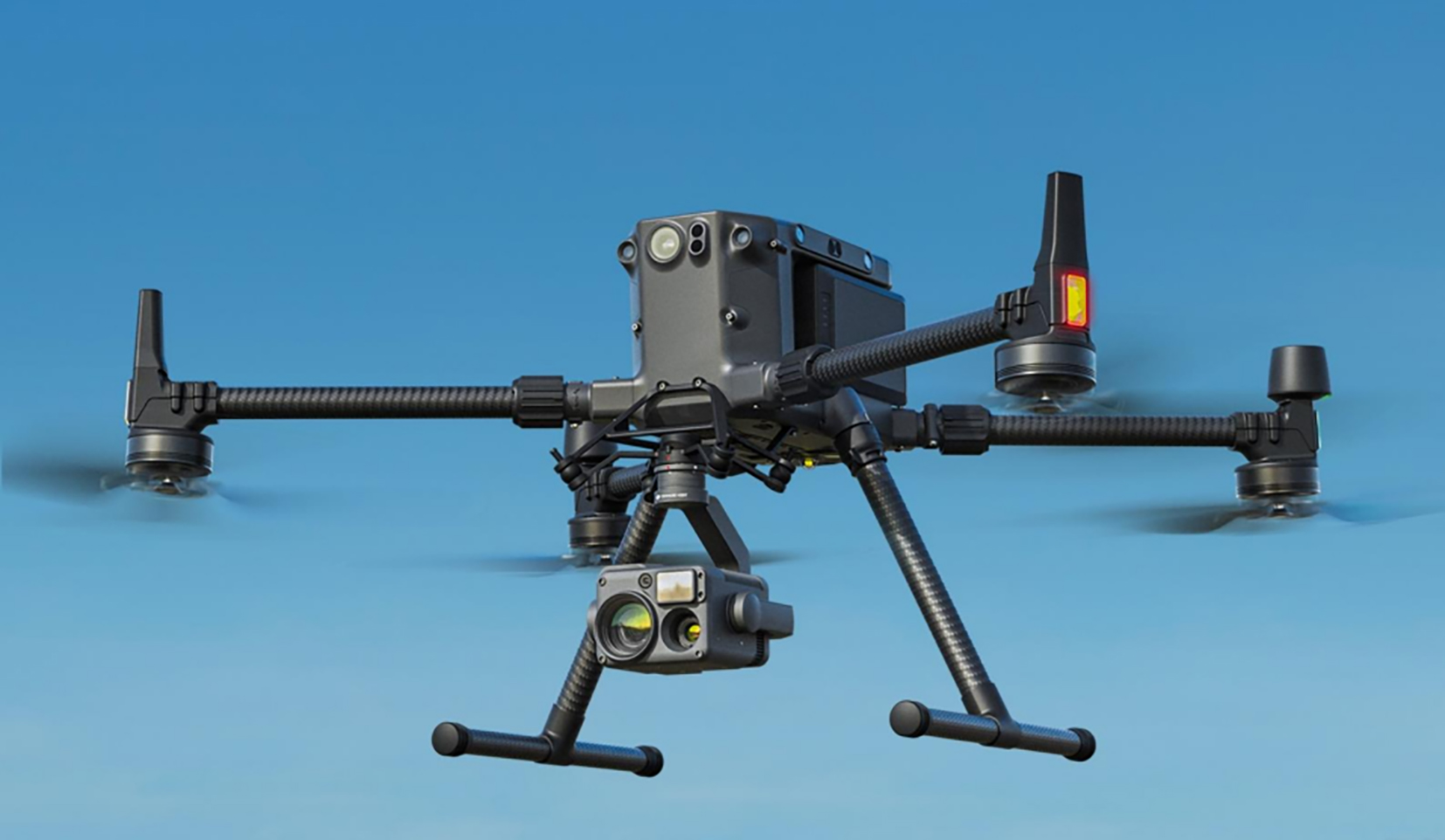}
    \caption{DJI Matrice 300 RTK UAV equipped with Zenmuse H20T gimbal camera used for infrared data acquisition in IRVL328.}
    \label{fig:irvl_uav}
\end{figure}

To address the lack of benchmark datasets for nighttime cross-modal localization, this study constructs a new infrared-visible dataset, termed IRVL328, on the campus of Nanjing University of Science and Technology, Nanjing, Jiangsu, China. The infrared images were collected using a DJI Matrice 300 RTK UAV equipped with a DJI Zenmuse H20T gimbal camera. The UAV platform provides a horizontal GPS accuracy of $\pm 1.5$\,m and a vertical GPS accuracy of $\pm 0.5$\,m. The thermal imaging camera of the Zenmuse H20T employs an uncooled Vanadium Oxide (\(\mathrm{VO}x\)) microbolometer sensor, with an image resolution of $640 \times 512$ and a frame rate of 30 frame/s. Its operating spectral range is $8$--$14$\,\si{\micro\metre}. The device supports 10 color palettes, and the white-hot palette is used in the experiments conducted in this study, as shown in Fig.~\ref{fig:irvl_uav}.

Data collection was conducted from April to June 2023 and covers diverse nighttime campus scenes. For the visible satellite modality, this study uses Gaofen-2 satellite imagery with a spatial resolution of 0.8\,m. The drone-satellite pairs are established by manual annotation, where each infrared drone image is matched with its corresponding visible-light satellite image according to the same geographic scene. In IRVL328, a scene is defined as a distinct cross-modal location pair corresponding to a unique geographic region on campus. In total, the dataset contains 328 such scenes and supports both drone-to-satellite nighttime localization and satellite-to-drone nighttime navigation. The dataset has been publicly released and is available at \url{https://drive.google.com/file/d/1okFeWJIuZ49TnkZkkoOYTOl_b0THwUdc/view?usp=sharing}.

\subsection{Evaluation metrics}

Three standard metrics are adopted for evaluation. Here, 
\(K_{\mathrm{ret}}\) denotes the number of top-ranked retrieved candidates 
considered in Recall@\(K_{\mathrm{ret}}\) evaluation, while 
\(k_{\mathrm{ret}}\) represents the rank index of an individual retrieved 
item in the ranked list. These symbols are independent of the variables 
used in the network formulation.

\noindent(1) Recall@\(K_{\mathrm{ret}}\) 
(R@\(K_{\mathrm{ret}}\)) evaluates whether a correct match is retrieved 
within the top-\(K_{\mathrm{ret}}\) results

\begin{equation}
\operatorname{Recall}@K_{\mathrm{ret}}
=
\frac{1}{|\mathcal{Q}|}
\sum_{q\in\mathcal{Q}}
\mathbb{I}
\big[
\operatorname{rank}(q)\leq K_{\mathrm{ret}}
\big]
\end{equation}

\noindent where \(\mathcal{Q}\) denotes the query set, 
\(\operatorname{rank}(q)\) represents the 1-based rank position of the 
first ground-truth match for query \(q\), and \(\mathbb{I}[\cdot]\) is 
the indicator function.

\noindent(2) Recall@$1$P (R@1P) measures whether a correct match is retrieved 
within the top \(1\%\) of the gallery

\begin{equation}
\operatorname{R@1P}
=
\frac{1}{|\mathcal{Q}|}
\sum_{q\in\mathcal{Q}}
\mathbb{I}
\big[
\operatorname{rank}(q)\leq K_{\mathrm{ret}}^{\mathrm{1P}}
\big]
\end{equation}

\begin{equation}
K_{\mathrm{ret}}^{\mathrm{1P}}
=
\max
\left(
1,
\left\lceil
0.01\,|\mathcal{G}|
\right\rceil
\right)
\end{equation}

\noindent where \(|\mathcal{G}|\) denotes the gallery size and 
\(K_{\mathrm{ret}}^{\mathrm{1P}}\) represents the number of retrieved 
candidates corresponding to the top \(1\%\) of the gallery, following 
the official evaluation protocol.

\noindent(3) Average Precision (AP) computes the area under the 
Precision--Recall curve. For a query \(q\), let \(\mathcal{G}_q\) denote 
the set of its ground-truth relevant items, and let 
\(y_q(k_{\mathrm{ret}})\in\{0,1\}\) indicate whether the retrieved item 
at the \(k_{\mathrm{ret}}\)th rank is relevant. Then

\begin{equation}
\left\{
\begin{aligned}
\operatorname{AP}(q)
&=
\frac{1}{|\mathcal{G}_q|}
\sum_{k_{\mathrm{ret}}=1}^{N}
P_q(k_{\mathrm{ret}})
y_q(k_{\mathrm{ret}})
\\[2pt]
P_q(k_{\mathrm{ret}})
&=
\frac{1}{k_{\mathrm{ret}}}
\sum_{i=1}^{k_{\mathrm{ret}}}
y_q(i)
\end{aligned}
\right.
\end{equation}

\noindent where \(N\) denotes the total number of ranked retrieved items 
considered for AP computation. The final AP is averaged over all queries.

\subsection{Implementation details and comparison protocol}
\label{V-C}

\textbf{Object Detection.} 
Two representative frameworks were evaluated: Faster R-CNN and YOLOv8. For Faster R-CNN, this study utilized a ResNet50 backbone with a Feature Pyramid Network (FPN) and a top-down attention mechanism. Parameters were initialized via Kaiming initialization,~\cite{ref66} with data augmentation including random cropping and horizontal flipping. Training was conducted for 40 epochs using SGD (momentum 0.9, weight decay 0.0001, batch size 8, learning rate 0.01). For YOLOv8, this study followed the official guidelines~\cite{ref47} and adopted YOLOv8n pre-trained weights. Training was performed for 200 epochs using SGD with momentum 0.937, weight decay 0.0005, batch size 32, and learning rate 0.01.

\textbf{Independent Detection Evaluation.}
Since object detection serves as the front-end for graph node construction, this study further conducts an independent detection evaluation to quantify the quality of the detected localization-relevant targets. Following the standard COCO-style evaluation protocol, this study reports $\mathrm{AP}_{50}$ and $\mathrm{AP}_{50:95}$ for all 16 categories used in this study. The results are summarized in Table~\ref{tab:det_map}. This evaluation is independent of the final image localization metrics and is used to verify whether the detector can provide reliable object-level observations for subsequent graph construction.

\begin{table}[t]
  \captionsetup{width=\linewidth}
  \centering
  \begin{threeparttable}
  \caption{Independent object detection performance for 16 localization-relevant categories.}
  \label{tab:det_map}
  \scriptsize
  \begin{tabular}{lcccc}
    \toprule
   \multirow{2}{*}{Category}
    & \multicolumn{2}{c}{Faster R-CNN}
    & \multicolumn{2}{c}{YOLOv8} \\
    \cmidrule(lr){2-3}\cmidrule(lr){4-5}
    & AP$_{50}$ (\%)
    & AP$_{50:95}$ (\%)
    & AP$_{50}$ (\%)
    & AP$_{50:95}$ (\%) \\
    \midrule
    Low-rise residential  & 59.97 & 34.94 & 48.69 & 24.99 \\
    Mid-rise residential  & 64.46 & 37.28 & 56.90 & 29.73 \\
    High-rise residential & 68.76 & 41.56 & 68.77 & 37.78 \\
    Saving box            & 54.16 & 27.95 & 32.37 & 18.62 \\
    Baseball field        & 70.63 & 44.06 & 62.57 & 31.52 \\
    Basketball field      & 73.84 & 42.66 & 64.09 & 33.97 \\
    Playground            & 84.35 & 52.16 & 70.31 & 41.18 \\
    Bridge                & 72.59 & 30.44 & 63.42 & 23.47 \\
    Irregular buildings   & 69.43 & 43.29 & 63.99 & 34.75 \\
    Intersection          & 57.27 & 24.07 & 42.89 & 15.30 \\
    Parking lot           & 63.45 & 28.58 & 54.80 & 21.67 \\
    Chimney               & 52.73 & 19.04 & 42.69 & 18.84 \\
    Tennis court          & 62.44 & 33.08 & 46.03 & 23.11 \\
    Football field        & 83.74 & 51.93 & 80.03 & 47.26 \\
    Rugby field           & 65.07 & 38.44 & 40.82 & 21.91 \\
    Lighthouse            & 60.89 & 36.07 & 27.55 & 13.74 \\
    \midrule
    Mean                  & 66.49 & 36.60 & 54.12 & 27.36 \\
    \bottomrule
  \end{tabular}

  \begin{tablenotes}
    \footnotesize
    \item Note: AP$_{50}$ and AP$_{50:95}$ are reported for Faster R-CNN and YOLOv8.
  \end{tablenotes}

  \end{threeparttable}
\end{table}

As shown in Table~\ref{tab:det_map}, Faster R-CNN achieves higher mean AP$_{50}$ and AP$_{50:95}$ than YOLOv8, indicating that it provides more reliable object-level observations for graph construction. Nevertheless, YOLOv8 offers a more lightweight alternative, which is further analyzed in the detector ablation experiments.

\textbf{Graph Neural Networks.} 
This study implemented a Siamese GNN architecture with TransformerConv as the primary operator. Drone and satellite graphs were processed by two shared TransformerConv layers: a 4-head attention layer for feature transformation and a subsequent layer for embedding fusion. Features were aggregated via global pooling and a fully connected classification head. Dropout with a rate of 0.5 was applied for regularization. Optimization was performed using AdamW with a learning rate of 0.0001 and a weight decay of $10^{-5}$.

\textbf{Baselines and Reproducibility.} 
This study reproduced several state-of-the-art baselines, including NetVLAD,~\cite{ref67} LPN,~\cite{ref7} University-1652,~\cite{ref19} FSRA,~\cite{ref23} MCCG,~\cite{ref30} and Sample4Geo.~\cite{ref32} Whenever available, official implementations were used; otherwise, the experimental settings reported in the original publications were followed as closely as possible. All experiments were conducted on an NVIDIA RTX 2080Ti GPU, and the random seed was fixed at 2024 to ensure reproducibility across all trials.

\textbf{Comparison Protocol.} 
 For the proposed method, the reported runtime includes the full inference pipeline, namely object detection, graph construction, graph reasoning, and final retrieval. Gallery/reference features were pre-extracted offline, and only query-side inference together with gallery matching was counted during testing, following common retrieval evaluation practice. For baseline methods, this study used their official implementations or default settings recommended by the original papers, and no additional acceleration tricks were introduced beyond those provided in the released code. 

\subsection{Quantitative evaluation}

To comprehensively evaluate the performance of the proposed model, this study conducted quantitative comparisons with classic methods in visual localization, such as NetVLAD,~\cite{ref67} as well as several recent advanced methods, including LPN,~\cite{ref7} University-1652,~\cite{ref19} FSRA,~\cite{ref23} MCCG,~\cite{ref30} and Sample4Geo.~\cite{ref32}

The experimental design adopts a strict cross-dataset evaluation strategy: the model is trained on the University-1652 dataset and tested on SUES-200, DenseUAV, and the self-collected infrared-visible dataset (IRVL328). This setting rigorously examines the model’s generalization, cross-modal adaptability, and robustness in unseen scenarios, thereby providing a comprehensive assessment for real-world applications.  

As summarized in Table~\ref{tab:table5}, the feasibility of this scheme arises from the structural consistency of the datasets. Nevertheless, a key challenge is the inherent data imbalance: each scene in University-1652, SUES-200, and IRVL328 typically includes multiple drone-view images but only one corresponding satellite-view image. Such imbalance may bias the learning process. To mitigate this issue, this study follows Ref.~\citenum{xiao2023long} and applies a threefold mining strategy on the satellite views, effectively reducing the negative impact of data imbalance on model training and evaluation.

Through experiments on SUES-200, IRVL328, and DenseUAV, this study summarizes the following key empirical findings:

\subsubsection{Competitive performance on SUES-200}

\begin{table*}[t]
  \captionsetup{justification=raggedright,singlelinecheck=false}
  \centering
  \caption{Comparison with state-of-the-art methods at different heights on SUES-200.}
  \label{tab:table2}

  \scriptsize
  \setlength{\tabcolsep}{2.5pt}
  \renewcommand{\arraystretch}{1.05}

  \begin{tabular}{c l *{10}{c}}
    \toprule

    \multirow{2}{*}{Height (m)}
    & \multirow{2}{*}{Method}
    & \multicolumn{5}{c}{Drone $\rightarrow$ Satellite}
    & \multicolumn{5}{c}{Satellite $\rightarrow$ Drone} \\

    \cmidrule(lr){3-7}
    \cmidrule(lr){8-12}

    & & R@1$\uparrow$ (\%)
    & R@5$\uparrow$ (\%)
    & R@10$\uparrow$ (\%)
    & R@1P$\uparrow$ (\%)
    & AP$\uparrow$ (\%)
    & R@1$\uparrow$ (\%)
    & R@5$\uparrow$ (\%)
    & R@10$\uparrow$ (\%)
    & R@1P$\uparrow$ (\%)
    & AP$\uparrow$ (\%) \\

    \midrule

    \multirow{8}{*}{150}
    & NetVLAD & 1.00 & 3.12 & 5.62 & 13.88 & 2.40
    & 1.25 & 1.25 & 7.50 & 52.50 & 1.04 \\

    & University-1652 & 23.62 & 52.12 & 66.38 & 41.38 & 30.44
    & 26.25 & 37.50 & 51.25 & 62.50 & 23.22 \\

    & LPN & 40.00 & 60.50 & 70.88 & 54.12 & 45.09
    & 45.09 & 45.09 & 76.25 & 78.75 & 40.84 \\

    & FSRA & 66.62 & 91.62$^{3}$ & 95.62$^{2}$ & 85.75$^{2}$ & 71.99
    & 70.00 & 83.75$^{3}$ & 85.00 & 92.50 & 61.73 \\

    & MuseNet & 68.62 & 85.88 & 90.12 & 81.62 & 72.47
    & 72.50 & 82.50 & 88.75$^{3}$ & 92.50 & 65.65 \\

    & MCCG & 72.38$^{3}$ & 90.75 & 95.12 & 85.12$^{3}$ & 76.22$^{3}$
    & 77.50$^{2}$ & 83.75$^{3}$ & 86.25 & 92.51$^{3}$ & 73.58$^{2}$ \\

    & Sample4Geo & 82.12$^{1}$ & 97.62$^{1}$ & 99.50$^{1}$ & 93.50$^{1}$ & 85.36$^{1}$
    & 88.75$^{1}$ & 95.00$^{1}$ & 97.50$^{1}$ & 98.75$^{1}$ & 85.81$^{1}$ \\

    & This study & 73.42$^{2}$ & 93.11$^{2}$ & 95.51$^{3}$ & 80.52 & 82.79$^{2}$
    & 74.59$^{3}$ & 85.29$^{2}$ & 89.71$^{2}$ & 98.53$^{2}$ & 72.94$^{3}$ \\

    \midrule

    \multirow{8}{*}{200}
    & NetVLAD & 1.12 & 3.25 & 5.38 & 12.12 & 2.40
    & 0.00 & 1.25 & 3.75 & 61.25 & 0.75 \\

    & University-1652 & 17.75 & 45.12 & 61.75 & 34.25 & 24.49
    & 20.00 & 28.75 & 43.75 & 53.75 & 16.93 \\

    & LPN & 25.25 & 46.62 & 57.62 & 40.12 & 30.75
    & 31.25 & 31.25 & 31.25 & 65.00 & 65.00$^{3}$ \\

    & FSRA & 52.88 & 83.25$^{2}$ & 89.50 & 76.25$^{3}$ & 59.63
    & 56.25 & 70.00 & 75.00 & 77.50 & 49.26 \\

    & MuseNet & 52.75 & 77.12 & 84.25 & 71.38 & 71.38$^{3}$
    & 60.00 & 60.00 & 81.25$^{3}$ & 81.25 & 52.90 \\

    & MCCG & 58.75$^{3}$ & 81.38 & 90.38$^{3}$ & 76.12 & 63.94
    & 61.25$^{3}$ & 76.25$^{3}$ & 80.00 & 90.00$^{2}$ & 57.08 \\

    & Sample4Geo & 77.25$^{1}$ & 94.12$^{1}$ & 98.12$^{1}$ & 89.50$^{1}$ & 80.74$^{2}$
    & 82.50$^{1}$ & 90.00$^{1}$ & 95.00$^{1}$ & 97.50$^{2}$ & 76.86$^{1}$ \\

    & This study & 73.97$^{2}$ & 91.67$^{2}$ & 96.08$^{2}$ & 87.97$^{2}$ & 78.35$^{2}$
    & 77.94$^{2}$ & 85.29$^{2}$ & 92.65$^{2}$ & 100.0$^{1}$ & 70.49$^{2}$ \\

    \midrule

    \multirow{8}{*}{250}
    & NetVLAD & 1.88 & 3.25 & 5.88 & 11.50 & 3.06
    & 0.00 & 2.50 & 7.50 & 56.25 & 0.98 \\

    & University-1652 & 26.12 & 54.75 & 67.25 & 44.00 & 32.83
    & 28.75 & 43.75 & 48.75 & 63.75 & 27.80 \\

    & LPN & 45.50 & 68.25 & 77.38 & 77.38 & 50.71
    & 60.00 & 71.25 & 76.25 & 86.25 & 86.25$^{2}$ \\

    & FSRA & 71.62 & 91.38 & 95.25 & 87.88$^{3}$ & 76.08
    & 73.75 & 83.75 & 90.00 & 91.25 & 65.68 \\

    & MuseNet & 76.12 & 76.12 & 91.88 & 84.25 & 78.81
    & 78.75 & 82.50 & 88.75 & 91.25 & 73.46 \\

    & MCCG & 80.00$^{3}$ & 94.12$^{3}$ & 96.62$^{3}$ & 90.75$^{2}$ & 83.07$^{3}$
    & 83.75$^{3}$ & 88.75$^{3}$ & 95.00$^{2}$ & 95.00$^{3}$ & 81.60 \\

    & Sample4Geo & 87.87$^{1}$ & 98.75$^{1}$ & 99.00$^{2}$ & 94.50$^{1}$ & 89.85$^{1}$
    & 93.75$^{1}$ & 96.25$^{1}$ & 97.50$^{1}$ & 98.75$^{1}$ & 88.87$^{1}$ \\

    & This study & 85.19$^{2}$ & 98.15$^{2}$ & 99.14$^{1}$ & 84.44 & 87.64$^{2}$
    & 89.41$^{2}$ & 92.65$^{2}$ & 94.11$^{3}$ & 97.06$^{2}$ & 85.38$^{3}$ \\

    \midrule

    \multirow{8}{*}{300}
    & NetVLAD & 0.88 & 3.88 & 7.12 & 12.75 & 2.30
    & 0.00 & 6.25 & 8.75 & 53.75 & 1.03 \\

    & University-1652 & 29.88 & 54.62 & 67.50 & 45.25 & 35.79
    & 35.00 & 43.75 & 50.00 & 66.25 & 29.86 \\

    & LPN & 54.62 & 72.62 & 81.25 & 81.25 & 58.88
    & 68.75 & 78.75 & 86.25 & 93.75 & 60.95 \\

    & FSRA & 74.38 & 92.38 & 95.62 & 88.50 & 78.34
    & 75.00 & 83.75 & 88.75 & 95.00 & 69.25 \\

    & MuseNet & 76.38 & 89.62 & 93.38 & 86.88 & 79.43
    & 77.50 & 77.50 & 86.25 & 86.25 & 86.25 \\

    & MCCG & 85.12$^{3}$ & 95.50$^{3}$ & 97.00$^{3}$ & 93.50$^{3}$ & 87.42$^{3}$
    & 88.75$^{3}$ & 90.00$^{3}$ & 93.75$^{3}$ & 96.25$^{3}$ & 85.72$^{3}$ \\

    & Sample4Geo & 89.87$^{1}$ & 97.75$^{2}$ & 99.75$^{2}$ & 95.50$^{2}$ & 91.54$^{1}$
    & 96.25$^{2}$ & 97.50$^{1}$ & 98.75$^{1}$ & 98.75$^{2}$ & 93.92$^{1}$ \\

    & This study & 88.08$^{2}$ & 98.81$^{1}$ & 99.90$^{1}$ & 95.99$^{1}$ & 87.95$^{2}$
    & 97.06$^{1}$ & 97.06$^{2}$ & 98.53$^{2}$ & 100.00$^{1}$ & 91.25$^{2}$ \\

    \bottomrule
  \end{tabular}

  \vspace{2pt}
  \begin{minipage}{\textwidth}
    \footnotesize
    Note: The top three results for each metric are shown, with the numbers 1, 2, and 3 indicating the rankings. The arrow $\uparrow$ denotes that higher values are better.
  \end{minipage}

\end{table*}

On the SUES-200 dataset, the proposed method achieves strong results across all altitudes (150–300 m) and evaluation metrics (R@1, R@5, R@10, R@1P, AP), consistently ranking among the top two (Table~\ref{tab:table2}). It performs particularly well on R@1 and R@10. For instance, at 300 m, the proposed method attains 88.08\% R@1 (vs 89.87\% for Sample4Geo) and at 250 m, 99.14\% R@10 (exceeding Sample4Geo’s 99.00\%). Although Sample4Geo slightly outperforms the proposed method on most metrics, the margin is typically within 1\%-2\%, highlighting the competitiveness of the proposed method. Moreover, the proposed method shows steadily improved performance with increasing altitude, confirming its robustness under challenging high-altitude conditions.

\subsubsection{Strong cross-modal transfer on IRVL328}

\begin{table*}[t]
  \captionsetup{width=\textwidth}
  \centering
  \caption{Comparison with state-of-the-art methods on the IRVL328 dataset.}
  \label{tab:table3}

  \scriptsize
  \setlength{\tabcolsep}{3pt}
  \renewcommand{\arraystretch}{1.05}

  \begin{tabular}{l *{10}{c}}
    \toprule

    \multirow{2}{*}{Method}
    & \multicolumn{5}{c}{Drone $\rightarrow$ Satellite}
    & \multicolumn{5}{c}{Satellite $\rightarrow$ Drone} \\

    \cmidrule(lr){2-6}
    \cmidrule(lr){7-11}

    & R@1$\uparrow$ (\%)
    & R@5$\uparrow$ (\%)
    & R@10$\uparrow$ (\%)
    & R@1P$\uparrow$ (\%)
    & AP$\uparrow$ (\%)
    & R@1$\uparrow$ (\%)
    & R@5$\uparrow$ (\%)
    & R@10$\uparrow$ (\%)
    & R@1P$\uparrow$ (\%)
    & AP$\uparrow$ (\%) \\

    \midrule

    NetVLAD
    & 0 & 1.73 & 1.73 & 9.83 & 0.83
    & 0.58 & 1.73 & 3.47 & 12.72 & 1.46 \\

    University-1652
    & 4.14 & 12.41 & 13.79 & 41.38$^{3}$ & 5.66
    & 4.38 & 12.16 & 15.83 & 10.89 & 6.64 \\

    LPN
    & 5.67 & 17.44 & 24.64 & 15.35 & 9.16
    & 23.45 & 40.00 & 46.21 & 77.24 & 7.72 \\

    FSRA
    & 13.21 & 29.91 & 39.27 & 27.41 & 17.73
    & 21.38 & 31.03 & 36.55 & 68.97 & 17.24 \\

    MuseNet
    & 18.75$^{3}$ & 39.79 & 53.05$^{3}$ & 36.89 & 24.10$^{3}$
    & 42.76$^{3}$ & 54.00$^{3}$ & 64.83$^{3}$ & 95.86$^{1}$ & 37.24$^{2}$ \\

    MCCG
    & 18.13 & 40.65$^{3}$ & 50.85 & 36.71 & 23.60
    & 38.62 & 51.86 & 61.38 & 91.03$^{3}$ & 20.78 \\

    Sample4Geo
    & 23.40$^{2}$ & 49.87$^{2}$ & 59.98$^{2}$ & 46.44$^{2}$ & 29.66$^{2}$
    & 50.34$^{1}$ & 62.07$^{2}$ & 66.90$^{2}$ & 90.34 & 29.75$^{3}$ \\

    This study
    & 36.09$^{1}$ & 58.22$^{1}$ & 66.68$^{1}$ & 55.84$^{1}$ & 41.47$^{1}$
    & 50.27$^{2}$ & 71.54$^{1}$ & 78.19$^{1}$ & 95.21$^{2}$ & 40.02$^{1}$ \\

    \bottomrule
  \end{tabular}

  \vspace{2pt}

  \begin{minipage}{\textwidth}
    \footnotesize
    Note: The top three results for each metric are shown, with the numbers 1, 2, and 3 indicating the rankings. The arrow $\uparrow$ denotes that higher values are better.
  \end{minipage}

\end{table*}

On IRVL328, the proposed method achieves the best performance on nearly all evaluation metrics, substantially outperforming competing methods (Table~\ref{tab:table3}). In particular, it reaches 36.09\% R@1 and 66.68\% R@10, clearly surpassing the strongest baseline, Sample4Geo, which attains 23.40\% and 59.98\%, respectively. These results highlight the superior cross-modal generalization capability of the proposed method under severe domain shift from visible-light training data (University-1652) to infrared testing data. The advantage mainly comes from its object-aware graph representation, which captures stable localization-relevant targets and explicitly models both spatial structure and semantic relations, thereby providing more robust matching cues than purely global feature-based methods. Although MuseNet also performs competitively, its performance is less consistent, whereas the proposed method maintains a clear and stable lead across almost all metrics.

\subsubsection{Robustness on DenseUAV}

\begin{table}[t]
  \captionsetup{width=\linewidth}
  \centering
  \caption{Comparison with state-of-the-art methods on DenseUAV dataset.}
  \label{tab:table4}

  \scriptsize
  \setlength{\tabcolsep}{3pt}
  \renewcommand{\arraystretch}{1.05}

  \begin{tabular}{l *{5}{c}}
    \toprule

    \multirow{2}{*}{Method}
    & \multicolumn{5}{c}{Drone $\rightarrow$ Satellite} \\

    \cmidrule(lr){2-6}

    & R@1$\uparrow$ (\%)
    & R@5$\uparrow$ (\%)
    & R@10$\uparrow$ (\%)
    & R@1P$\uparrow$ (\%)
    & AP$\uparrow$ (\%) \\

    \midrule

    NetVLAD
    & 0.09 & 0.64 & 1.59 & 36.68 & 0.22 \\

    University-1652
    & 3.99 & 11.80 & 16.69 & 47.40 & 2.95 \\

    LPN
    & 5.62 & 12.40 & 17.29 & 52.90 & 3.05 \\

    FSRA
    & 17.72 & 39.68 & 50.19 & 38.15 & 15.85 \\

    MuseNet
    & 13.13 & 31.45 & 41.96 & 80.22 & 8.35 \\

    MCCG
    & 23.68$^{3}$ & 50.66$^{3}$ & 61.18$^{3}$ & 84.39$^{3}$ & 84.60$^{1}$ \\

    Sample4Geo
    & 26.42$^{2}$ & 54.22$^{2}$ & 66.58$^{2}$ & 95.36$^{1}$ & 19.27$^{3}$ \\

    This study
    & 50.34$^{1}$ & 83.98$^{1}$ & 88.44$^{1}$ & 86.99$^{2}$ & 76.59$^{2}$ \\

    \bottomrule
  \end{tabular}

  \vspace{2pt}

  \begin{minipage}{\linewidth}
    \footnotesize
    Note: The top three results for each metric are shown, with the numbers 1, 2, and 3 indicating the rankings. The arrow $\uparrow$ denotes that higher values are better.
  \end{minipage}

\end{table}

On DenseUAV, which is characterized by highly complex scenes, dense background clutter, and pronounced spatiotemporal variations, the proposed method achieves the best performance in the drone-to-satellite task (Table~\ref{tab:table4}). It attains 50.34\% R@1 and 88.44\% R@10, significantly outperforming Sample4Geo, which achieves 26.42\% and 66.58\%, respectively. These results demonstrate that the proposed method remains robust under challenging scene complexity and substantial temporal and spatial appearance changes, and is better able to capture stable localization-relevant structures across views.

\subsubsection{Cross-dataset generalization discussion}
In comparison with Sample4Geo, while it slightly outperforms the proposed method on the SUES-200 dataset, the proposed method demonstrates superior cross-modal and complex-scene adaptability on IRVL328 and DenseUAV. Similarly, MCCG and FSRA show competitiveness on certain metrics but lack overall robustness. In contrast, baseline methods such as NetVLAD and University-1652 exhibit poor performance across all evaluated datasets, confirming their limitations in handling significant cross-view and cross-modal variations.

Nevertheless, the current experiments still cannot fully establish the universality of the proposed method across all remote sensing scenarios. Future work will further validate the proposed framework on broader geographic regions, additional sensor types, and larger-scale cross-domain benchmarks.

\subsection{Ablation study}

To further validate the proposed framework, this study conducts ablation studies from five perspectives: the necessity of graph modeling, the contribution of semantic and spatial graphs, the role of global and local information, the contribution of different training objectives, and the impact of the spatial weighting strategy. All experiments are conducted on IRVL328, which contains diverse nighttime cross-modal scenes with substantial variations in viewpoint, illumination, and thermal radiation. The results are reported in Table~\ref{tab:table6}.

\begin{table*}[t]
  \captionsetup{width=\linewidth}
  \centering
  \caption{Ablation study on IRVL328.}
  \label{tab:table6}

  \scriptsize
  \setlength{\tabcolsep}{3pt}
  \renewcommand{\arraystretch}{1.05}

  \begin{tabular}{l *{10}{c}}
    \toprule

    \multirow{2}{*}{Model}
    & \multicolumn{5}{c}{Drone $\rightarrow$ Satellite}
    & \multicolumn{5}{c}{Satellite $\rightarrow$ Drone} \\

    \cmidrule(lr){2-6}
    \cmidrule(lr){7-11}

    & R@1$\uparrow$ (\%)
    & R@5$\uparrow$ (\%)
    & R@10$\uparrow$ (\%)
    & R@1P$\uparrow$ (\%)
    & AP$\uparrow$ (\%)
    & R@1$\uparrow$ (\%)
    & R@5$\uparrow$ (\%)
    & R@10$\uparrow$ (\%)
    & R@1P$\uparrow$ (\%)
    & AP$\uparrow$ (\%) \\

    \midrule

    Object-only w/o graph
    & 21.24 & 35.56 & 50.35 & 54.45 & 33.05
    & 30.21 & 37.46 & 54.13 & 47.24 & 34.41 \\

    Graph w/o message passing
    & 68.27 & 76.00 & 86.19 & 88.96 & 57.69
    & 71.05 & 75.89 & 81.08 & 94.66 & 45.99 \\

    w/o $\bm{V}_{\mathrm{sp}}$
    & 81.38 & 95.17 & 96.55 & 98.31 & 47.56
    & 83.45 & 95.86 & 97.24 & 99.31 & 60.19 \\

    w/o $\bm{V}_{\mathrm{se}}$
    & 62.23 & 78.36 & 89.45 & 97.68 & 65.82
    & 35.37 & 55.32 & 65.95 & 91.50 & 14.04 \\

    w/o Global
    & 75.53 & 80.24 & 93.82 & 98.23 & 70.15
    & 78.13 & 92.52 & 94.83 & 98.55 & 50.22 \\

    w/o $L_{\mathrm{node}}$
    & 70.53 & 85.27 & 90.42 & 97.15 & 60.34
    & 72.66 & 88.11 & 92.48 & 97.39 & 40.22 \\

    w/o $L_{\mathrm{M}}$
    & 53.42 & 68.67 & 77.09 & 56.28 & 47.46
    & 35.11 & 53.99 & 60.11 & 80.16 & 30.45 \\

    w/o $L_{\mathrm{cls}}$
    & 67.43 & 93.65 & 97.61 & 91.28 & 73.00
    & 80.67 & 93.79 & 95.17 & 99.31 & 38.77 \\

    w/o $\cos(\cdot)$
    & 82.25 & 98.30 & 99.46 & 97.69 & 85.83
    & 75.64 & 85.32 & 95.96 & 92.02 & 54.05 \\

    Full model
    & 86.19 & 99.01 & 99.68 & 98.47 & 89.03
    & 91.03 & 97.93 & 98.62 & 100.00 & 63.65 \\

    \bottomrule
  \end{tabular}

  \vspace{2pt}

  \begin{minipage}{\textwidth}
    \footnotesize
    Note: The table reports the performance of different model variants for both Drone $\rightarrow$ Satellite and Satellite $\rightarrow$ Drone retrieval. The arrow $\uparrow$ denotes that higher values are better.
  \end{minipage}

\end{table*}

\subsubsection{Necessity of graph modeling}
This study first examines whether explicit graph modeling is necessary for cross-modal geo-localization beyond simply detecting salient objects. To this end, this study introduces two additional baselines. The first, Object-only w/o graph, removes graph construction entirely and directly aggregates detector-derived object features using pooling and an MLP-based matching head. The second, Graph w/o message passing, preserves the same graph construction as the full model but disables message passing, such that node features are aggregated without relational propagation.

These two variants are designed to disentangle the contributions of object detection, graph construction, and graph reasoning. As shown in Table~\ref{tab:table6}, Object-only w/o graph yields the weakest performance in both retrieval directions, indicating that object-level cues alone are insufficient for robust cross-view matching. Introducing graph construction without message passing already brings a substantial improvement, which suggests that object-aware graph representation provides a useful structural prior. However, the full model still achieves the best overall performance, with R@1 improving from 68.27\% to 86.19\% for Drone$\rightarrow$Satellite and from 71.05\% to 91.03\% for Satellite$\rightarrow$Drone. These results demonstrate that the performance gain comes not only from object detection or graph construction, but more importantly from explicit relational reasoning over graph-structured object representations.

\subsubsection{Contribution of semantic and spatial graphs}
The proposed framework jointly models a semantic graph (\(\bm{V}_{\mathrm{se}}\)) and a spatial graph (\(\bm{V}_{\mathrm{sp}}\)) to capture complementary information for cross-modal matching. To evaluate their individual contributions, this study removes each graph component in turn.

When the spatial graph is removed, the model still achieves a relatively high R@1 of 81.38\% in the Drone$\rightarrow$Satellite direction, indicating that semantic cues provide strong discriminative information for top-ranked retrieval. However, the AP drops sharply from 89.03\% to 47.56\%, showing that spatial relations are important for improving ranking quality and retrieval stability. In contrast, removing the semantic graph leads to a much larger degradation in the main retrieval metrics. In particular, R@1 decreases to 62.23\% for Drone$\rightarrow$Satellite and further drops to 35.37\% for Satellite$\rightarrow$Drone, revealing that semantic information serves as the dominant cue for cross-modal correspondence. Overall, these results indicate that semantic and spatial graphs play complementary roles: the semantic graph provides the primary discriminative signal, while the spatial graph introduces structural constraints that improve robustness and ranking consistency. The best performance is achieved when both graphs are jointly modeled.

\subsubsection{Contribution of global and local information}
In the proposed framework, a scene-level global node is introduced and fused with local object nodes to construct the matching graph. This design aims to complement local object evidence with holistic scene context, especially when object detection is incomplete or unstable under cross-modal appearance changes. To evaluate the contribution of global information, this study removes the global node and retains only local object-level graph matching.

Without the global node, R@1 decreases from 86.19\% to 75.53\% in the Drone$\rightarrow$Satellite direction and from 91.03\% to 78.13\% in the Satellite$\rightarrow$Drone direction. The AP also drops markedly, from 89.03\% to 70.15\% and from 63.65\% to 50.22\%, respectively. These results confirm that global context substantially improves both retrieval accuracy and ranking stability. They also support the design motivation that global information provides complementary scene-level guidance when local object relations alone are insufficient for reliable matching.

\subsubsection{Contribution of training objectives}
The proposed model is trained with three objectives: graph node matching loss (\(L_{\mathrm{node}}\)), graph embedding matching loss (\(L_{\mathrm{M}}\)), and graph classification loss (\(L_{\mathrm{cls}}\)). To analyze their respective roles, this study removes each objective individually during training.

Removing \(L_{\mathrm{node}}\) leads to a clear performance drop, with R@1 decreasing to 70.53\% for Drone$\rightarrow$Satellite and 72.66\% for Satellite \\ $\rightarrow$Drone. This indicates that node-level supervision is important for learning fine-grained cross-modal correspondences between graph nodes. 

Removing \(L_{\mathrm{M}}\) causes the most severe degradation, with R@1 dropping to 53.42\% and 35.11\% in the two retrieval directions, respectively, showing that this term serves as the primary retrieval objective and is indispensable for aligning graph-level embeddings. 

Removing \(L_{\mathrm{cls}}\) also leads to notable degradation, particularly in terms of R@1, where the performance drops to 67.43\% for Drone$\rightarrow$Satellite and 80.67\% for Satellite$\rightarrow$Drone. This suggests that the classification objective provides important auxiliary scene-level supervision and helps improve feature discrimination and optimization stability, rather than serving as the dominant matching signal. 

Overall, the three objectives play complementary roles: \(L_{\mathrm{M}}\) drives global retrieval alignment, \(L_{\mathrm{node}}\) strengthens local correspondence learning, and \(L_{\mathrm{cls}}\) improves discriminative consistency at the scene level.

\subsubsection{Impact of spatial weighting strategy}
\label{sec:ablation_cosine}
To construct the spatial graph, this study incorporates cosine similarity $\cos(\cdot)$ into the edge weights so that both geometric relations and feature consistency are considered. To evaluate the effect of this design, this study removes the cosine weighting term and retains only the geometric component.

As shown in Table~\ref{tab:table6}, removing cosine weighting causes a consistent performance drop in both retrieval directions. In the Drone$\rightarrow$Satellite direction, R@1 decreases from 86.19\% to 82.25\% and AP decreases from 89.03\% to 85.83\%. The degradation is even more evident in the Satellite$\rightarrow$Drone direction, where R@1 drops from 91.03\% to 75.64\% and AP falls from 63.65\% to 54.05\%. These results indicate that feature-aware edge weighting improves the reliability of spatial relation modeling, particularly under stronger cross-modal discrepancies. Therefore, cosine weighting is not a dominant source of improvement by itself, but it consistently enhances the robustness and effectiveness of the spatial graph.

Overall, the ablation results in Table~\ref{tab:table6} demonstrate that the proposed framework benefits from the coordinated interaction of graph modeling, semantic-spatial graph design, global-local feature integration, and multi-objective optimization. Removing any key component leads to an overall degradation across most retrieval metrics.

\subsection{Comparative experiments}
To identify the optimal network architecture, this study further conducts comparative experiments by varying the model backbone, object detection head, graph embedding pooling methods, and training strategies. Similar to the ablation studies, all comparative experiments are carried out on the self-collected IRVL328 infrared-visible dataset.

\subsubsection{Selection of model backbone}

\begin{table*}[t]
  \captionsetup{width=\linewidth}
  \centering
  \caption{Impact of Graph Neural Network modules on final localization performance.}
  \label{tab:table7}

  \scriptsize
  \setlength{\tabcolsep}{3pt}
  \renewcommand{\arraystretch}{1.05}

  \begin{tabular}{l *{10}{c}}
    \toprule

    \multirow{2}{*}{Model}
    & \multicolumn{5}{c}{Drone $\rightarrow$ Satellite}
    & \multicolumn{5}{c}{Satellite $\rightarrow$ Drone} \\

    \cmidrule(lr){2-6}
    \cmidrule(lr){7-11}

    & R@1$\uparrow$ (\%)
    & R@5$\uparrow$ (\%)
    & R@10$\uparrow$ (\%)
    & R@1P$\uparrow$ (\%)
    & AP$\uparrow$ (\%)
    & R@1$\uparrow$ (\%)
    & R@5$\uparrow$ (\%)
    & R@10$\uparrow$ (\%)
    & R@1P$\uparrow$ (\%)
    & AP$\uparrow$ (\%) \\

    \midrule

    GCN
    & 75.10 & 96.26 & 98.39 & 95.14 & 79.71
    & 73.10 & 89.66 & 94.48 & 98.62 & 52.91 \\

    GraphSAGE
    & 81.07 & 97.96 & 99.48 & 97.10 & 84.83
    & 75.17 & 95.86 & 97.24 & 99.31 & 56.20 \\

    GAT
    & 82.26 & 98.36 & 99.46 & 99.44 & 85.84
    & 83.45 & 96.55 & 97.24 & 99.31 & 60.17 \\

    TransformerConv
    & 86.19 & 99.01 & 99.68 & 98.47 & 89.03
    & 91.03 & 97.93 & 98.62 & 100.00 & 63.65 \\

    \bottomrule
  \end{tabular}

  \vspace{2pt}

  \begin{minipage}{\textwidth}
    \footnotesize
    Note: The arrow $\uparrow$ indicates that higher values are better.
  \end{minipage}

\end{table*}

As shown in Table~\ref{tab:table7}, which compares different graph neural network modules, the overall performance gap among the models is relatively small. Nevertheless, TransformerConv consistently achieves the best results across multiple metrics, particularly in R@1 and AP, demonstrating its superior ability to model relationships between graph nodes. By contrast, GCN exhibits the weakest performance, reflecting its limited capacity to capture complex spatial and semantic dependencies. GraphSAGE and GAT achieve more balanced results across metrics but still fall short of TransformerConv. Therefore, this study adopts TransformerConv as the backbone to enhance matching accuracy.

\subsubsection{Selection of object detection head}

\begin{table*}[t]
  \captionsetup{width=\linewidth}
  \centering
  \caption{Comparison of object detection heads.}
  \label{tab:table8}

  \scriptsize
  \setlength{\tabcolsep}{3pt}
  \renewcommand{\arraystretch}{1.05}

  \begin{tabular}{l *{10}{c}}
    \toprule

    \multirow{2}{*}{Model}
    & \multicolumn{5}{c}{Drone $\rightarrow$ Satellite}
    & \multicolumn{5}{c}{Satellite $\rightarrow$ Drone} \\

    \cmidrule(lr){2-6}
    \cmidrule(lr){7-11}

    & R@1$\uparrow$ (\%)
    & R@5$\uparrow$ (\%)
    & R@10$\uparrow$ (\%)
    & R@1P$\uparrow$ (\%)
    & AP$\uparrow$ (\%)
    & R@1$\uparrow$ (\%)
    & R@5$\uparrow$ (\%)
    & R@10$\uparrow$ (\%)
    & R@1P$\uparrow$ (\%)
    & AP$\uparrow$ (\%) \\

    \midrule

    Faster R-CNN
    & 86.19 & 99.01 & 99.68 & 98.47 & 89.03
    & 91.03 & 97.93 & 98.62 & 100.00 & 63.65 \\

    YOLOv8
    & 76.14 & 80.99 & 89.40 & 80.99 & 53.68
    & 53.25 & 74.68 & 83.44 & 97.40 & 26.50 \\

    \bottomrule
  \end{tabular}

  \vspace{2pt}

  \begin{minipage}{\textwidth}
    \footnotesize
    Note: The arrow $\uparrow$ indicates that higher values are better.
  \end{minipage}

\end{table*}

Table~\ref{tab:table8} compares the performance of Faster R-CNN and YOLOv8 in the matching task. Faster R-CNN achieves superior results across all recall and average precision metrics, particularly in the Drone$\rightarrow$Satellite direction for R@1 (86.19\% vs 76.14\%) and AP (89.03\% vs 53.68\%), demonstrating its advantage in providing more accurate object detection and thus enhancing matching performance. In contrast, YOLOv8 shows clear strengths in inference speed, making it suitable for real-time applications. Nevertheless, for scenarios where accuracy is prioritized, Faster R-CNN remains the preferred choice.

\subsubsection{Impact of pooling methods}

\begin{table*}[t]
  \captionsetup{width=\linewidth}
  \centering
  \caption{Comparison of feature aggregation methods.}
  \label{tab:table9}

  \scriptsize
  \setlength{\tabcolsep}{3pt}
  \renewcommand{\arraystretch}{1.05}

  \begin{tabular}{l *{10}{c}}
    \toprule

    \multirow{2}{*}{Model}
    & \multicolumn{5}{c}{Drone $\rightarrow$ Satellite}
    & \multicolumn{5}{c}{Satellite $\rightarrow$ Drone} \\

    \cmidrule(lr){2-6}
    \cmidrule(lr){7-11}

    & R@1$\uparrow$ (\%)
    & R@5$\uparrow$ (\%)
    & R@10$\uparrow$ (\%)
    & R@1P$\uparrow$ (\%)
    & AP$\uparrow$ (\%)
    & R@1$\uparrow$ (\%)
    & R@5$\uparrow$ (\%)
    & R@10$\uparrow$ (\%)
    & R@1P$\uparrow$ (\%)
    & AP$\uparrow$ (\%) \\

    \midrule

    GRU
    & 82.24 & 97.32 & 98.42 & 95.70 & 84.83
    & 83.45 & 94.86 & 97.24 & 98.31 & 54.19 \\

    Mean Pooling
    & 83.25 & 98.33 & 99.45 & 97.70 & 85.83
    & 84.14 & 94.48 & 97.93 & 100.00 & 63.65 \\

    GEM
    & 86.19 & 99.01 & 99.68 & 98.47 & 89.03
    & 91.03 & 97.93 & 98.62 & 100.00 & 63.65 \\

    \bottomrule
  \end{tabular}

  \vspace{2pt}

  \begin{minipage}{\textwidth}
    \footnotesize
    Note: The arrow $\uparrow$ indicates that higher values are better.
  \end{minipage}

\end{table*}

Table~\ref{tab:table9} reports the results of different feature aggregation methods. Adaptive pooling (GEM) achieves the highest performance across all metrics, with R@1 reaching 86.19\% and AP reaching 89.03\% in the Drone$\rightarrow$Satellite direction. In contrast, average pooling and GRU yield slightly lower results. These findings suggest that GEM is more effective in capturing global features, thereby improving overall matching accuracy.

\subsubsection{Dynamic weight update during training}

\begin{table*}[t]
  \captionsetup{width=\linewidth}
  \centering
  \caption{Impact of adaptive dynamic weight updates on model performance.}
  \label{tab:table10}

  \scriptsize
  \setlength{\tabcolsep}{3pt}
  \renewcommand{\arraystretch}{1.05}

  \begin{tabular}{l *{10}{c}}
    \toprule

    \multirow{2}{*}{Model}
    & \multicolumn{5}{c}{Drone $\rightarrow$ Satellite}
    & \multicolumn{5}{c}{Satellite $\rightarrow$ Drone} \\

    \cmidrule(lr){2-6}
    \cmidrule(lr){7-11}

    & R@1$\uparrow$ (\%)
    & R@5$\uparrow$ (\%)
    & R@10$\uparrow$ (\%)
    & R@1P$\uparrow$ (\%)
    & AP$\uparrow$ (\%)
    & R@1$\uparrow$ (\%)
    & R@5$\uparrow$ (\%)
    & R@10$\uparrow$ (\%)
    & R@1P$\uparrow$ (\%)
    & AP$\uparrow$ (\%) \\

    \midrule

    No weight update
    & 76.09 & 96.41 & 98.66 & 95.32 & 80.64
    & 81.38 & 93.10 & 97.24 & 99.31 & 55.94 \\

    Weight update
    & 86.19 & 99.01 & 99.68 & 98.47 & 89.03
    & 91.03 & 97.93 & 98.62 & 100.00 & 63.65 \\

    \bottomrule
  \end{tabular}

  \vspace{2pt}

  \begin{minipage}{\textwidth}
    \footnotesize
    Note: The arrow $\uparrow$ indicates that higher values are better.
  \end{minipage}

\end{table*}

As shown in Table~\ref{tab:table10}, models with dynamically updated weights in multi-task learning outperform those with fixed weights across all metrics. For example, in the Drone$\rightarrow$Satellite direction, R@1 increases from 76.09\% to 86.19\%, while AP improves from 80.64\% to 89.03\%. These results demonstrate that adaptive weight adjustment during training significantly enhances both generalization and matching accuracy.

In summary, this study adopts TransformerConv as the backbone for its superior performance in key metrics, Faster R-CNN for more accurate object detection, and GEM pooling for effective global feature extraction. Moreover, dynamic weight updates further improve accuracy and generalization, yielding an optimized balance between accuracy and efficiency in the matching task.

\subsection{Computational complexity, runtime fairness, and scalability}
\label{subsec:efficiency}

\subsubsection{Theoretical complexity}
The computational cost of the proposed drone-satellite image matching framework mainly comes from three stages: object detection, graph construction, and graph reasoning.

\textbf{Object Detection.}
For Faster R-CNN, the backbone feature extraction has complexity approximately proportional to the input resolution, i.e., \(O(WHC)\), where \(W \times H\) denotes the image size and \(C\) is the channel number. The Region Proposal Network further generates candidate regions, and the subsequent IoU computation and non-maximum suppression introduce a cost dominated by \(O(P^2)\), where \(P\) is the number of proposals retained for post-processing. ROI pooling and region classification add an additional \(O(PF)\), where \(F\) is the region feature dimension. In practice, since the number of retained candidate regions is constrained to a small range, object detection remains tractable but still constitutes the dominant computational bottleneck.

\textbf{Graph Construction.}
Let \(K_{\mathrm{node}}\) denote the number of valid graph nodes in one image, including the global node and all detected object nodes. In the proposed framework, \(K_{\mathrm{node}}\) is explicitly bounded (\(K_{\mathrm{node}} \leq 30\) in practice). The spatial graph requires pairwise geometric computation between detected regions, while the semantic graph computes pairwise affinities between node features. Therefore, graph construction has complexity \(O(K_{\mathrm{node}}^2)\). Because \(K_{\mathrm{node}}\) is small, this cost is negligible compared with object detection.

\textbf{Graph Reasoning.}
For a GNN layer with feature dimension \(F\), the computational cost can be written as \(O(EF + K_{\mathrm{node}}F^2)\), where \(E\) is the number of edges. In the proposed framework, the semantic graph is fully connected and the spatial graph is sparse, so the overall number of edges remains moderate due to the small number of nodes. With two TransformerConv layers, the graph reasoning stage remains lightweight in practice. Since \(K_{\mathrm{node}}\) is bounded, the GNN computation is effectively a low-order overhead compared with the detector.

Overall, the dominant runtime cost comes from object detection, while graph construction and graph reasoning introduce only limited additional overhead due to the bounded number of graph nodes. This design makes the proposed method computationally practical while still benefiting from explicit relation modeling.

\subsubsection{Runtime fairness protocol}

\begin{table*}[t]
  \captionsetup{width=\linewidth}
  \centering
  \caption{Runtime comparison on the SUES-200 test set under identical hardware conditions.}
  \label{tab:Runtime}

  \scriptsize
  \setlength{\tabcolsep}{4pt}
  \renewcommand{\arraystretch}{1.05}

  \begin{tabular}{c c *{8}{c}}
    \toprule

    & & \multicolumn{8}{c}{Runtime (s)} \\
    \cmidrule(lr){3-10}

    Height (m)
    & Direction
    & NetVLAD
    & University-1652
    & LPN
    & FSRA
    & MuseNet
    & MCCG
    & Sample4Geo
    & This study \\

    \midrule

    \multirow{2}{*}{150}
    & D2S
    & 20.81 & 9.80 & 15.09 & 11.02 & 13.20 & 10.35 & 10.17 & 6.08 \\

    & S2D
    & 78.84 & 14.77 & 26.84 & 18.08 & 18.74 & 16.84 & 18.56 & 10.28 \\

    \midrule

    \multirow{2}{*}{200}
    & D2S
    & 21.32 & 9.82 & 14.75 & 10.97 & 13.65 & 10.29 & 10.87 & 6.43 \\

    & S2D
    & 79.02 & 14.58 & 25.20 & 17.99 & 19.36 & 16.84 & 18.82 & 10.41 \\

    \midrule

    \multirow{2}{*}{250}
    & D2S
    & 21.41 & 9.76 & 13.79 & 11.12 & 10.96 & 10.35 & 10.83 & 6.41 \\

    & S2D
    & 80.05 & 15.10 & 24.17 & 17.84 & 17.94 & 16.74 & 17.28 & 10.64 \\

    \midrule

    \multirow{2}{*}{300}
    & D2S
    & 21.21 & 9.35 & 13.70 & 11.00 & 14.66 & 10.33 & 10.83 & 6.42 \\

    & S2D
    & 81.45 & 14.52 & 24.21 & 17.90 & 17.96 & 16.92 & 17.26 & 9.63 \\

    \bottomrule
  \end{tabular}

  \vspace{2pt}

  \begin{minipage}{\textwidth}
    \footnotesize
    Note: D2S denotes drone-to-satellite retrieval, and S2D denotes satellite-to-drone retrieval. For the proposed method, the reported runtime includes the complete online query-side pipeline, including object detection, graph construction, graph reasoning, and final retrieval matching, while gallery/reference features are pre-extracted offline following standard retrieval practice.
  \end{minipage}

\end{table*}

To ensure fair efficiency comparison, the runtime results reported in Table~\ref{tab:Runtime } are measured under the same hardware environment. All methods are tested on an NVIDIA RTX 2080Ti GPU. For the proposed method, the reported runtime includes the complete online inference pipeline on the query side, including object detection, graph construction, graph reasoning, and final retrieval matching. Preprocessing and model loading are not counted. Following standard retrieval practice, gallery/reference features are pre-extracted offline and reused during testing. The same evaluation principle is applied to baseline methods whenever their official implementations support feature precomputation. Official implementations or the default settings reported in the original papers are used for the compared methods, and no additional acceleration tricks are introduced beyond those available in the released code. This protocol is intended to minimize implementation bias and make the runtime comparison as consistent as possible.

Under this protocol, OGMN achieves the fastest runtime among the compared methods on the SUES-200 standard test set. In the D2S task, the proposed method requires only 6.08--6.43\,s across different altitudes, corresponding to a 69.7\%--70.8\% latency reduction relative to NetVLAD. In the more challenging S2D task, where the gallery size increases from 200 to 10000 images, OGMN requires 9.63--10.64\,s, yielding an 86.7\%--88.2\% latency reduction compared with NetVLAD. These results indicate that the proposed framework remains highly efficient even when detector-based graph construction is included in the online pipeline. Moreover, the runtime variation across different flight heights remains small, indicating that the method is not overly sensitive to altitude changes in practical deployment.

\subsubsection{Single-query efficiency under different input resolutions}

To further evaluate the practical efficiency and deployment potential of the proposed framework, this study conducts a detailed single-query analysis under different input resolutions. Specifically, this study evaluates \(224 \times 224\), \(384 \times 384\), and \(512 \times 512\) input sizes under the query drone-to-gallery satellite protocol. The query number is 3129 and the gallery size is 166. The results are reported in Table~\ref{tab:single_query_efficiency}.

\begin{table*}[t]
  \captionsetup{width=\linewidth}
  \centering
  \caption{Single-query efficiency analysis under different input resolutions on RTX 2080Ti.}
  \label{tab:single_query_efficiency}
  \scriptsize

  \resizebox{\linewidth}{!}{
  \begin{tabular}{lcccccccccc}
    \toprule
    Input size &
    Latency &
    Latency P50 &
    Latency P95 &
    Throughput &
    Detection &
    Graph matching &
    Peak GPU &
    GPU increment &
    Avg. nodes &
    Accuracy variation \\

    &
    (ms) &
    (ms) &
    (ms) &
    (QPS) &
    (\%) &
    (\%) &
    (GB) &
    (GB) &
    &
    (\%) \\

    \midrule

    \(224 \times 224\)
    & 13.33
    & 13.18
    & 14.34
    & 75.03
    & 71.14
    & 7.98
    & 1.76
    & 0.28
    & 10.23
    & Reference \\

    \(384 \times 384\)
    & 15.81
    & 15.60
    & 16.95
    & 63.23
    & 60.70
    & 7.11
    & 1.77
    & 0.28
    & 13.24
    & \(-1.52\) to \(+3.19\) \\

    \(512 \times 512\)
    & 19.65
    & 18.74
    & 21.71
    & 50.90
    & 51.53
    & 6.70
    & 1.77
    & 0.28
    & 14.56
    & \(-22.86\) to \(-7.39\) \\

    \bottomrule
  \end{tabular}
  }

  \vspace{2pt}
  \footnotesize
  Note: Latency denotes the average single-query latency. P50 and P95 denote the median and 95th percentile latency, respectively. Detection ratio denotes the proportion of object detection time in the total single-query latency. Graph matching ratio includes graph construction, GNN reasoning, and retrieval matching. Accuracy variation reports the relative change compared with the \(224 \times 224\) setting, without listing the original accuracy values.

\end{table*}

As shown in Table~\ref{tab:single_query_efficiency}, the proposed method maintains efficient single-query inference under different input resolutions. When the input size increases from \(224 \times 224\) to \(512 \times 512\), the average single-query latency increases from 13.33\,ms to 19.65\,ms, while the throughput decreases from 75.03 queries per second to 50.90 queries per second. This indicates that higher input resolutions introduce additional computational overhead, but the overall query-side inference speed remains efficient.

The stage-wise runtime analysis further shows that object detection is the dominant computational component. Across the three input resolutions, the object detection stage accounts for 51.53\%--71.14\% of the total single-query latency. In contrast, the graph matching stage, including graph construction, GNN reasoning, and retrieval matching, accounts for only 6.70\%--7.98\%. These results confirm that the proposed object-aware graph matching module itself is lightweight, and that the major computational cost mainly comes from the detection front-end rather than the graph reasoning module.

In terms of memory usage, the peak GPU memory remains stable at approximately 1.76--1.77\,GB across different input resolutions, while the additional GPU memory increment remains around 0.28\,GB. This suggests that the proposed graph-based matching pipeline has stable memory consumption and is not sensitive to moderate changes in input resolution. The average number of graph nodes increases from 10.23 to 14.56 as the input resolution increases, but the graph-related computation remains bounded because the maximum number of valid nodes is explicitly limited.

Regarding localization accuracy, the \(384 \times 384\) setting maintains a stable performance trend compared with the \(224 \times 224\) setting, with relative changes across evaluation metrics ranging from \(\mathord{-}1.52\%\) to \(\mathord{+}3.19\%\). When the resolution is further increased to \(512 \times 512\), the retrieval performance decreases by approximately 7.39\%--22.86\% across different evaluation metrics. This result indicates that simply increasing the input resolution does not necessarily improve cross-domain localization performance. A possible reason is that higher resolution may introduce more detected regions, larger scale variation, and more complex graph structures, which can increase the difficulty of stable cross-view graph matching. Therefore, \(224 \times 224\) and \(384 \times 384\) provide a better trade-off between localization accuracy and computational efficiency.

\subsubsection{Scalability analysis}
The proposed framework is also designed with large-gallery retrieval in mind. First, gallery/reference embeddings can be extracted offline and stored in advance, so online inference mainly consists of query-side object detection, graph construction, graph reasoning, and similarity computation against the precomputed gallery embeddings. As a result, when the gallery size increases, the main additional online cost lies in the final similarity matching stage rather than repeated graph construction for gallery images.

Second, the graph size is explicitly controlled by limiting the number of valid detected targets. This keeps both graph construction and GNN reasoning bounded, preventing the cost from growing excessively with scene complexity. In other words, the graph-related overhead is constrained mainly by the detector output cap, rather than by the raw image resolution or the total number of possible regions.

Third, from a storage and memory perspective, each gallery image is represented by a compact graph-level embedding after GEM aggregation, rather than by the full set of intermediate graph structures. This substantially reduces the memory burden during retrieval and makes large-scale matching feasible. Therefore, although the detector remains the main online bottleneck, the proposed framework scales favorably to larger galleries because most gallery-side computation is performed offline, while query-side graph reasoning remains lightweight due to the bounded node number.

Finally, this study acknowledges that the current experiments are conducted on a desktop GPU rather than an embedded platform. Therefore, this study describes the proposed method as having potential for onboard deployment rather than claiming real-time onboard deployment. Future work will further validate the proposed framework on embedded platforms such as Jetson AGX Orin or Jetson Xavier, and will investigate lightweight detector replacement, model pruning, and quantization to improve practical deployment efficiency.

\subsection{Visualization and analysis}

To provide intuitive insight into the behavior of the proposed method, this study visualizes representative retrieval results and intermediate attention responses. These visualizations help illustrate both the strengths and the remaining limitations of the proposed framework.

\begin{figure}[t]
    \centering
    \includegraphics[width=\linewidth]{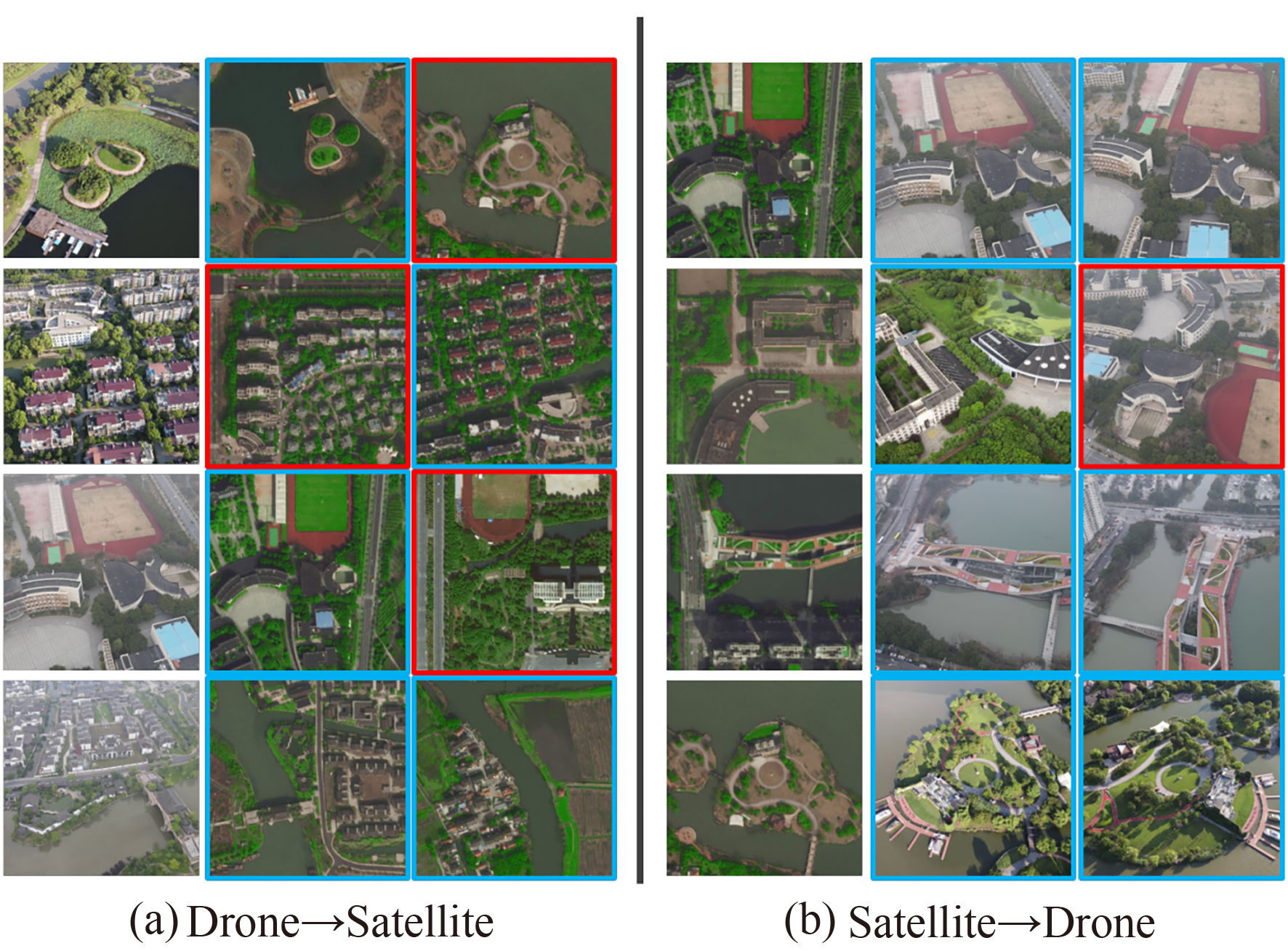}
    \caption{Qualitative retrieval results of the proposed method.}
    \label{fig5}
\end{figure}

\begin{figure}[t]
    \centering
    \includegraphics[width=\linewidth]{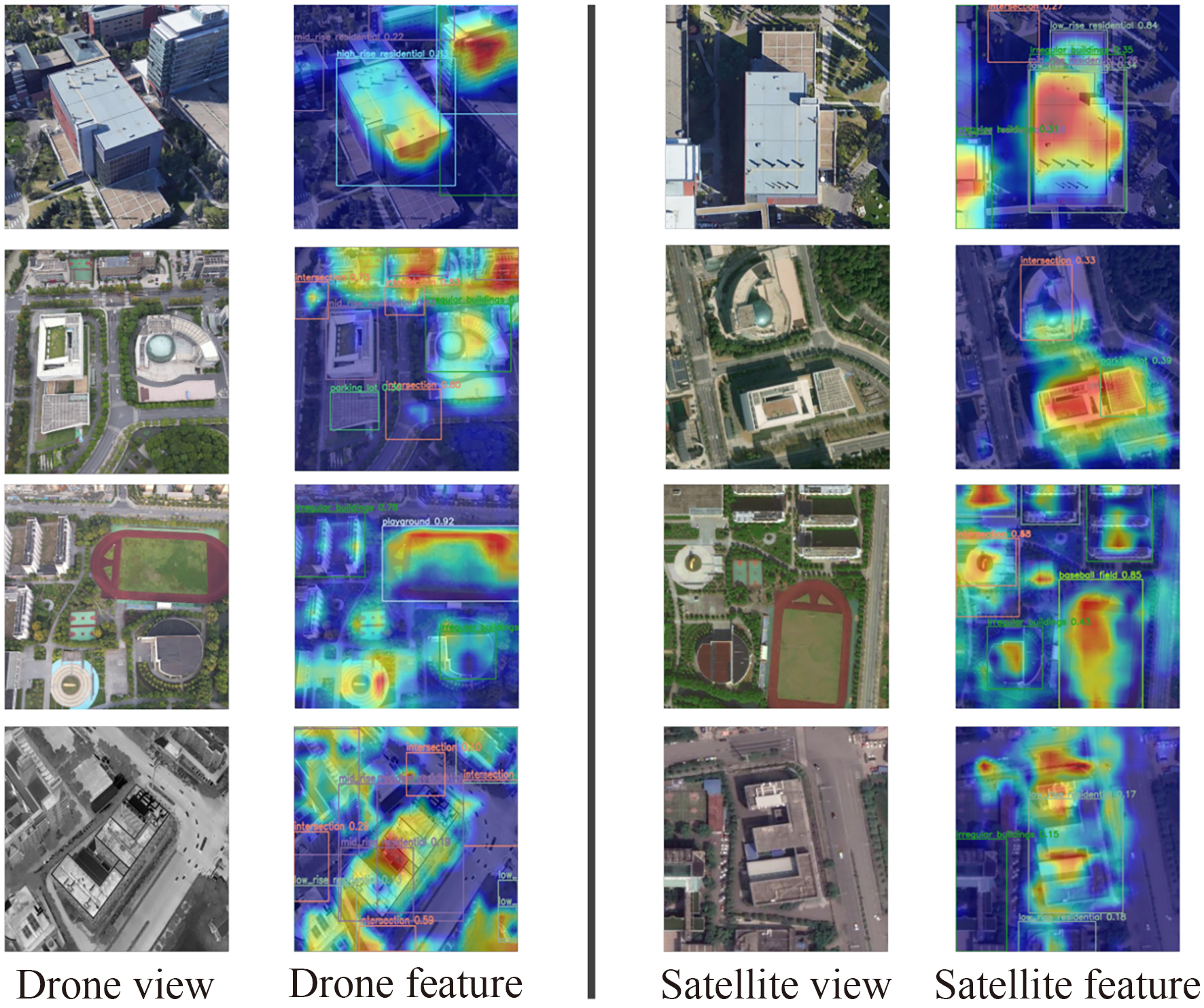}
    \caption{Visualization of intermediate attention responses in proposed method.}
    \label{fig6}
\end{figure}

\subsubsection{Visualization of matching results}
Fig.~\ref{fig5} presents qualitative retrieval results of the proposed method, including both successful and failed cases for drone-view object localization (a) and drone navigation (b). The top two retrieved candidates are displayed from left to right according to their confidence scores, where blue boxes indicate correct matches and red boxes denote incorrect matches. In most cases, the top-ranked retrieval results correspond to the ground-truth targets, demonstrating that the proposed framework can effectively establish cross-view correspondences between drone and satellite images. 

Nevertheless, several failure cases are also observed. Although these incorrectly retrieved candidates often share similar visual appearances or structural patterns with the query images, they do not correspond to the actual target regions. These failures suggest that the learned representation is capable of capturing meaningful scene-level and object-level cues, while still facing challenges caused by appearance similarity, viewpoint variations, and structural ambiguities in complex cross-view scenarios.

\subsubsection{Heatmap analysis}
This study further employs Grad-CAM \cite{ref69} to visualize the intermediate responses of the model, as illustrated in Fig.~\ref{fig6}. The resulting heatmaps show that the model focuses on semantically meaningful and spatially stable regions, such as buildings, roads, and other prominent structures. In addition, the attention responses remain relatively concentrated on viewpoint-invariant regions across different image modalities, suggesting that the proposed framework effectively integrates semantic and spatial cues.

\section{Conclusions}
\label{sec:conclusions}

This study proposed OGMN for cross-source remote sensing image localization. The main conclusions are summarized as follows:
\begin{enumerate}[label=(\arabic*), leftmargin=*, itemsep=0pt, topsep=2pt, parsep=0pt]
    \item OGMN integrates object detection with graph neural networks to model spatial topology and semantic relations, improving feature alignment across drone and satellite views.
    \item Experiments on University-1652, SUES-200, DenseUAV, and IRVL328 show that OGMN achieves competitive performance against strong baselines, and generalizes well under temporal shifts and cross-modal discrepancies.
    \item OGMN attains favorable accuracy-efficiency trade-offs, with reduced inference latency while maintaining robust localization performance, supporting its potential for onboard deployment.
    \item The proposed graph-based formulation offers an interpretable modeling alternative for multi-source remote sensing image matching.
    \item Future work will explore online/adaptive learning to further enhance robustness in dynamic real-world scenarios.
\end{enumerate}

\section*{Acknowledgements}

This work was supported by the National Natural Science Foundation of China 
(No. 62175111) and the Open Research Fund of Jiangsu Key Laboratory of Visual 
Sensing and Intelligent Perception, China (No. JSSJCG202601).

\bibliographystyle{cja-word-custom}
\bibliography{cja-template-word-style}

\end{document}